\documentclass{article}

\usepackage{arxiv}

\usepackage[utf8]{inputenc} 
\usepackage[T1]{fontenc}    

\usepackage{hyperref}       
\usepackage{url}            
\usepackage{booktabs}       
\usepackage{amsfonts}       
\usepackage{nicefrac}       
\usepackage{microtype}      
\usepackage{lipsum}
\usepackage{graphicx}
\usepackage{amssymb, amsmath}
\title{Deep Learning for Change Detection in Remote Sensing Images: Comprehensive Review and  Meta-Analysis}

\author{
  Lazhar khelifi\thanks{webpage: https://www.lazharkhelifi.com} \\
Department of Computer Science\\ and Operations Research \\
Montreal University\\
Montreal, Quebec, Canada \\
\texttt{khelifil@iro.umontreal.ca} \\
   \And
 Max Mignotte\\
  Department of Computer Science\\ and Operations Research \\
  Montreal University\\
  Montreal, Quebec, Canada \\
  \texttt{mignotte@iro.umontreal.ca} \\
}

\begin{document}
\maketitle

\begin{abstract}
Deep learning (DL) algorithms are considered as a methodology of
choice for remote-sensing image analysis over the past few years.  Due
to its effective applications, deep learning has also been introduced
for automatic change detection and achieved great success.  The
present study  attempts to provide a comprehensive review and a
meta-analysis of the recent progress in this subfield. Specifically,
we first introduce the fundamentals of deep learning methods which are
frequently adopted for change detection. Secondly, we present the
details of the meta-analysis conducted to examine the status of change
detection DL studies. Then, we focus on deep learning-based change
detection methodologies for remote sensing images by giving a general
overview of the existing methods. Specifically, these deep
learning-based methods were classified  into three groups; fully
supervised learning-based methods, fully unsupervised learning-based
methods and transfer learning-based techniques. As a result of these
investigations, promising new directions were identified for future
research. This study will contribute in several ways to our
understanding of deep learning for change detection and will provide a
basis for further research.
\end{abstract}


\section{Introduction}
\label{introduction}

Deep learning (DL) has seen an increasing trend and a great interest
over the past decade due to its powerful ability to represent
learning. Deep learning allows models that are built, based on
multiple processing layers, to learn representations of data samples
with several levels of abstraction \cite{LeCun2015}.  Deep learning
enables models that are composed, based on multiple layers, to learn
representations of data samples with several ranges of abstraction
levels \cite{LeCun2015}. It may also be considered as the analysis of
models that either require a greater composition of learned concepts
or functions, compared to conventional machine learning models such as
naive Bayes \cite{rish2001empirical} \cite{Lewis1998}, support vector
machine (SVM) \cite{Suykens99leastsquares} \cite{CauwenberghsP00},
random forests, \cite{breiman2001random} \cite{GISLASON2006294}  and
the decision tree\cite{Safavian1991}\cite{FRIEDL1997}.

\vspace*{0.2em}
On the basis of its state-of-the-art performance, deep learning 
has been consequently applied to various domains, such as computer vision \cite{Voulodimos}, 
speech recognition \cite{6639344}, and information retrieval \cite{Palangi2016}. Particularly, in the computer vision field, deep
learning has taken great leaps thanks to the recent advances of
processing power, the improvements in graphics processors  and the
increased data volumes (i.e., videos and images).  Notably, the
science of remote sensing (RS) has seen a massive increase in the
generation and enhancement of digital images captured from airplanes
or satellites that cover almost each angle of the surface of the earth.
This growth in data has pushed the community of the geoscience and remote sensing (RS) to apply deep learning algorithms to solve
different remote sensing tasks. Among these tasks, stands out the
change detection (CD) task defined in \cite{SINGH1989} as '\textit{the
	process of identifying differences in the state of an object or
	phenomenon by observing it at different times}'. In another word,
change detection refers to identifying the differences between images
acquired over the same geographical zone but taken at two distinct times \cite{GONG2017212}.

\vspace*{0.2em}
Change detection techniques are extensively utilized  in various applications \cite{9037317} including; disaster assessment \cite{Bovolo2007}, environmental  monitoring \cite{Coppin2004}, land management \cite{FERANEC2007234} and  urban  change analysis \cite{VIANA2019621},
etc. Currently, the number of extreme disasters caused by climate
change such as drought, floods, hurricanes, and heat waves, has
revealed at the same time a new challenge for researchers and a need
for developing more effective automated change detection
methods. Motivated by those aforementioned observations, deep learning
has been introduced for change detection in remote sensing and
achieved good performance.

\vspace*{0.2em}
Recently, various reviews that focus on deep learning for remote
sensing data have been published. These studies have summarized the deep learning techniques adopted in all major remote sensing sub-areas
including classification, restoration,
denoising, target recognition, scene understanding, and other tasks
(for further details we refer the reader to \cite{MA2019166} \cite{8113128} \cite{7486259}). 
To the best of our knowledge, however, there is no work that has studied the recent progress of deep learning for the task of change detection in a specific and extensive way.
Therefore, the purpose of this present report is to provide an overview of the state of deep learning algorithms as applied in remote sensing images for change detection.
Hence, by performing a meta-analysis, we selected and
categorized the relevant papers related to DL and change detection. By
doing so, then we provide a technical review of these studies that
shed more light on the advance of deep learning for change
detection. This review will serve as a base for future studies in this
subfield of research.

\vspace*{0.2em}
The rest of this paper is structured as follows.  Section
\ref{Change_detection} presents the definition of the change detection
problem. Section \ref{Overviewdeep} gives a brief overview of deep
learning as well as the typical deep models used for change
detection. Section \ref{Methods_data} describes the methods and data
used to review the state-of-the-art. In Section \ref{Deep_learning},
we divide these previous works into three categories; fully supervised
learning-based methods, fully unsupervised learning-based methods, and
transfer learning-based methods. Section \ref{Promising_research}
suggests two interesting  research directions to further advance the
field. Finally, Section \ref{Conclusion} outlines the conclusions.

%
%
\section{Change detection in remote sensing} 
\label{Change_detection}

\begin{figure*}[!t]
	\centering
	\includegraphics[width=14cm]{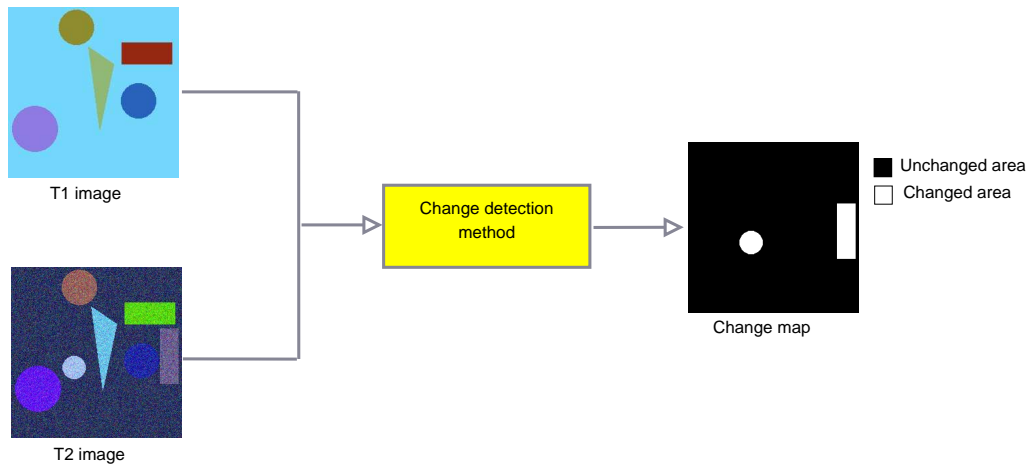}
	\caption{\label{Fig1}  Graphical illustration of the change detection problem.}
\end{figure*}

\begin{figure*}[!t]
	\centering
	\includegraphics[width=14cm]{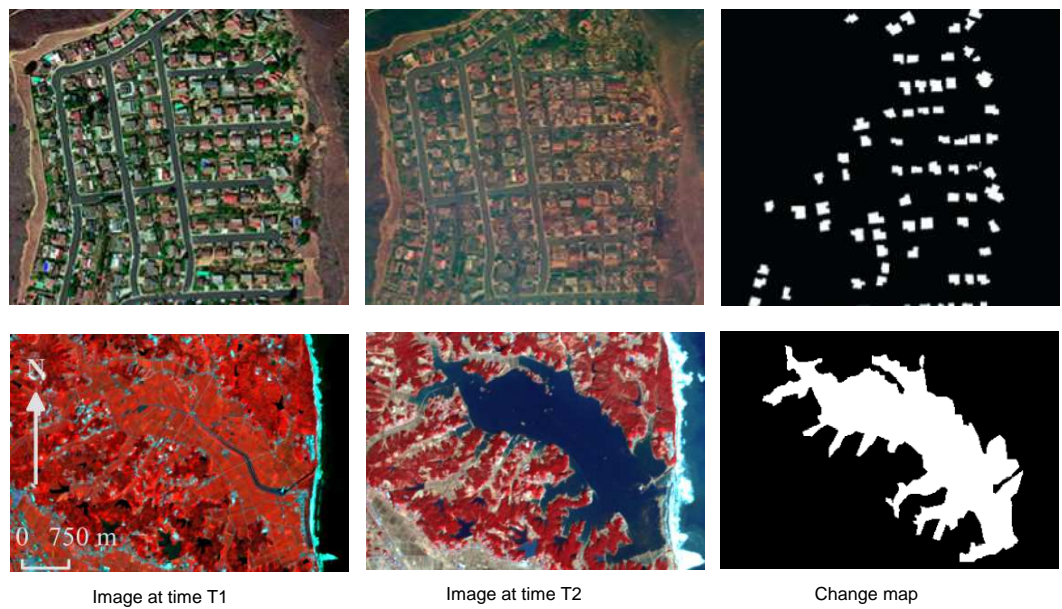}
	\caption{\label{Fig2}
		An illustration of a typical change detection results
		within a high-resolution satellite image \cite{Sublime_2019} \cite{Kolos2019}.}
\end{figure*}

Change detection is the operation of quantitative analysis and determination of surface changes from phenomena or objects over two distinct periods \cite{SINGH1989}.
This process, which is a basic technology in the field of earth observation, attempts to distinguish the changed and unchanged pixels of bi-temporal or multi-temporal remote sensing images acquired from the same geographical zone or area, but at different times, respectively \cite{Liu2015} \cite{Yang2019_bb}.
Assigning to each pixel a binary label based on a pair or series of co-registered images represents the main purpose of the change detection system. A positive label thus means that the area of that pixel has changed, while a null label represents an unchanged area (See Figs. \ref{Fig1} and \ref{Fig2}) \cite{Liu2019}.
Actually, change detection represents a powerful tool for video surveillance, mapping urban areas, and other forms of multi-temporal analysis.

Formally, let $\mathcal{I}_1$ and $\mathcal{I}_2$  be two co-registered images, which share
the same size $W$ $\times$ $L$ and taken over the same geographical
region at two separate periods $t_1$ and $t_2$, respectively, using the
same sensor, in the classic monomodal case:
%
\begin{equation}
\label{E2}            
\mbox{ $\mathcal{I}_1$ }= \{ \mathcal{I}_1(x, y), 1 \leq x \leq W, 1 \leq  y \leq L\}
\end{equation}
\begin{equation}
\label{I2a} 
\mbox{ $\mathcal{I}_2$ }= \{ \mathcal{I}_2(x, y), 1 \leq x \leq W, 1 \leq y \leq L\}.
\end{equation}
%
The primary purpose of a change detection system is to generate an accurate binary change map (CM):
%
\begin{equation}
\label{I2b} 
\mbox{CM}= \{ CM(x,y) \in  \{0,1\} , 1 \leq x \leq W, 1 \leq y \leq L\},  
\end{equation}
%
where  $(x,y)$ represents the position coordinates of the pixel
indexed $i$.  In traditional methods, this change map can be obtained
by a difference image (DI) operation, based on differencing or
log-rationing function ($DI=|\mathcal{I}_1 - \mathcal{I}_2|$), followed by a final analysis
of the DI result. 

Change detection  has been successfully used in a wide variety of
applications. In particular, in the agricultural sector, change
detection is adopted  for deforestation monitoring, disaster
assessment and  shifting cultivation monitoring. In the military
field, it is now utilized in collecting information about new military
installations, movement of the enemy's military forces, battlefield
area, and damage assessment \cite{Kevin2018}. In the civil field,
change detection is used to control urban area development and city
extension \cite{Kadhim2016}. Also, it is actually adopted to monitor
the effects of climate changes usually associated with the increase of levels of greenhouse gas (GHG) emissions in the atmosphere, such as changes in mass balance and glacier facies or sea-level change.

While the change detection algorithms have shown many benefits in
various fields of applications, it faces some serious
challenges. Among these challenges, we can consider the variation in
data acquisition parameters which can affect the process of finding
the relevant changes by adding an irrelevant information into the
data. In addition, this unwanted change  can be emerged as atmospheric
features, like fog, clouds, and dust. For example, a cloud present in
one image (at time $t1$) but not in the other one (at time $t2$) leads
to a bright patch that can be registered as a difference and
consequently affects the quality of the resulting change map.
Angles of sunlight may also present
another problem related to the presence and the direction of the shadows on the scene \cite{Kevin2018}. 
Besides, vegetation growth and surface reflectance of objects
such as soil before and after rain can also affect the result of a
change \cite{ZhangDianjun2016}. Thus, a robust change detection method must be able to differentiate between relevant changes and irrelevant changes in satellite images in addition to the detection of temporal changes.
Motivated by those successful
applications, recently deep learning techniques, capable of extracting information from data (image or video), have been applied to solve
this problem and have achieved good performances.

%
%

%
%
\section{Brief Overview of deep learning}
\label{Overviewdeep}

Deep learning (DL) algorithms, aiming at learning representative and discriminatory features from a set of data in a hierarchical way, have  received much attention from worldwide geoscience
and remote sensing communities, in recent years. In the first part of this section, we briefly present the deep learning history to explain the trend in its growth. In the second part, we outline different deep network models
widely designed for change detection in remote sensing
images. These deep networks incorporate deep belief networks (DBNs),
stacked autoencoders (SAEs), generative adversarial networks
(GANs), recurrent neural networks (RNNs), and convolutional neural networks (CNNs).

\subsection{History}
\label{history}

Deep learning (DL) is a particular approach of machine learning which
takes advantage of the knowledge of the statistics, human brain, and applied mathematics  statistics, as it advanced over the last years \cite{Goodfellow2016}. By gathering these pieces of knowledge, this approach  relieved human experts  from formally defining all the knowledge that the computer machine requires to resolve a particular problem. This powerful approach reaches good  flexibility and scalability by representing the world as an embedded hierarchical structure of
concepts. This concept hierarchy  enables the machine to recognize  complex concepts by developing them from simpler ones \cite{Goodfellow2016}.
DL is driven from the connectivity theory related to the functionality of our brain cells, also called neurons, leading to the concept of artificial neural networks (ANN). ANN is designed based on artificial neuron layers to receive input data  and
transform it into outputs by applying an activation function and
learning progressively higher-level features. The intermediate layers
(in the middle of the input and output) are often called ``\textit{hidden
	layers}'' because they are not directly observable from the inputs
and outputs of the system \cite{reed99a}.
In practice, to solve complex tasks such as change detection in remote
sensing images, a neural network that contains multiple hidden layers
is applied. This multiple-layered structure is addressed as a
``\textit{deep}'' neural networks (DNNs), therefore, the word
``\textit{deep learning}''.

As described in \cite{Goodfellow2016}, the development of deep
learning has followed three main waves reflecting different
philosophical viewpoints. The first wave refers to cybernetics in the
1940s-1960s, characterized by the concepts advance  of biological learning \cite{McCullochPitts43} \cite{hebb:behavior} and
application of the first models such as the perceptron
\cite{rosenblatt1958perceptron} which allows the training of an
architecture based on a unique neuron. The second stage began with
the connectionist\footnotemark[1] approach expanded in  the 1980s-1995s period,
with back-propagation \cite{Rumelhart:1986we} to train a neural
network using one or two hidden layers. 
This fundamental building block updates the weights of the connections in the network for multiple times, by
minimizing a measure of the gap among the actual output vector of
the net and the aimed output vector \cite{Rumelhart1988}. While this
approach works quite well when dealing with simple applications,
especially, the community of computer vision has found some issues to
apply this approach to complex problems. The main challenge was the
lack of specific computing hardware to train efficiently deep neural
networks (DNNs).  The third wave started in 2006 under the name of
deep learning \cite{Hinton:2006} \cite{Bengio:LeCun:07}. Since that
time, 
we have seen a renewed importance in deep neural networks benefitted to the availability of  powerful computer systems, expanded databases and new training techniques. 
Currently, deep learning has
received much focus in different research areas of computer
vision, including the analysis of remote sensing images.

\subsection{Deep models}
\label{Deep_Models}

\subsubsection{DBNs}
\label{DBNs}

\begin{figure}[!t]
	\centering
	\includegraphics[width=8cm]{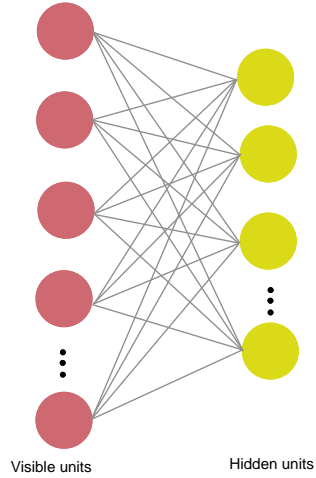}
	\caption{\label{rbm} A general illustration of RBM.}
\end{figure}

\begin{figure*}[!t]
	\centering
	\includegraphics[width=12cm]{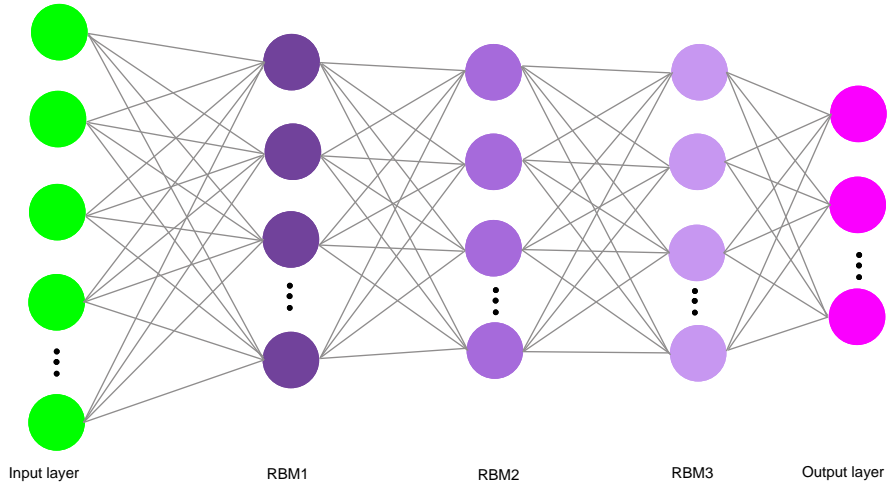}
	\caption{\label{DBN}General illustration of DBN.}
\end{figure*}

Deep belief networks (DBNs) are mainly built based on a layerwise
training model called restricted Boltzmann machine (RBM). RBMs are
stochastic undirected graphical models containing a layer of visible
variables and a unique layer of hidden variables. Fig. \ref{rbm}
illustrates the graph structure of the RBM.  
It is a bipartite graph that involves the link of  visible units representing observations, to hidden units that learn to describe features based on undirected weighted connections \cite{Mohamed5704567}. 
In this model, there are
no connections permitted among any variables in the visible layer or
between any units in the hidden layer. Mathematically, let suppose
that the visible layer $v$ contains a set of $n_{v}$ binary random
variables, and the hidden layer $h$ consists of $n_{h}$ binary random
variables. The energy function of the canonical RBM can be formulated
as \cite{Goodfellow2016}:
%
\begin{equation}
\label{E} 
\mbox{E(v,h)} = - b^\top v - c^\top h - v^\top  W h
\end{equation}
%
where $W$, $b$ and $c$ represent learnable parameters. The weights $W$
on the connections and the biases $b$  
The weights on the connexions and the biases of the individual units express a distribution of probability through an energy function over the joint states of the visible and hidden units \cite{Mohamed5947494}. The probability (i.e., energy) of a joint configuration is then defined as:
%
\begin{equation}
\label{p} 
\mbox{p(v,h)} = Z \exp (-E(v,h)) 
\end{equation}
%
where $Z$ is the normalizing constant usually referred to  the partition function:
%
\begin{equation}
\label{Z} 
\mbox{Z}  =   \sum_{v} \sum_{h} \exp \left\{ - E(v,h) \right\}
\end{equation}
%
Because of the restricted characteristic (i.e., feature) representation capability of  a unique RBM, several RBMs can be stacked one by one forming a DBN that may effectively trained  to obtain a deep hierarchical modeling of the training data \cite{Li8697135}. Fig. \ref{DBN} presents a DBN
composed by stacking multiple RBM layers.

%
%

%
%
\footnotetext[1]{Connectionism represents  a movement in cognitive science that aims to interpret intellectual abilities through artificial neural networks \cite{Buckner}.} 
%

\subsubsection{SAEs}
\label{SAEs}

\begin{figure*}[!t]
	\centering
	\includegraphics[width=8.5cm]{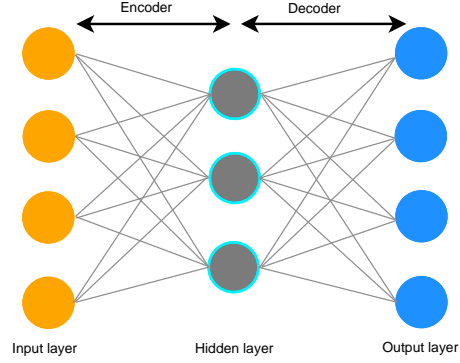}
	\caption{\label{ea}Auto-Encoders.}
\end{figure*}

\begin{figure*}[!t]
	\centering
	\includegraphics[width=8.5cm]{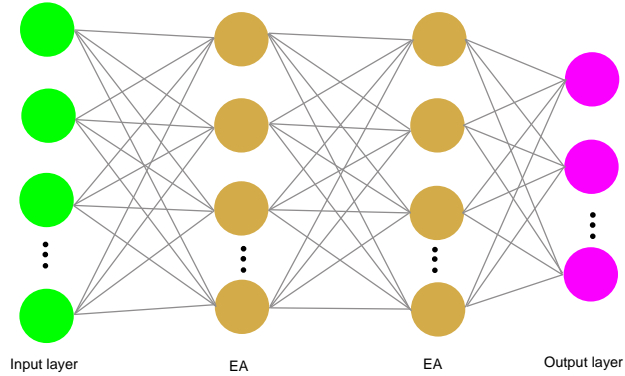}
	\caption{\label{sea}Stacked Auto-Encoders.}
\end{figure*}

Autoencoder (AE) is considered as the principal building piece of the
stacked autoencoder (SAE) \cite{ZABALZA20161}. An autoencoder is a
feedforward neural network model that applies backpropagation, setting the objective values to be consistent (or equal) to the inputs. This model consists of two
steps an encoder $h = f(x)$ and a decoder that attempts to provide a
reconstruction $r = g(h)$. On the one hand, based on a non-linear function, the encoder side projects the input vector $(x)$ to the hidden layer:
%
\begin{equation}
\label{h} 
\mbox{h} = s(w_{h} x + b_{h}) 
\end{equation}
%
On the other hand, the decoder maps the hidden layer back to the output layer that contains an identical number of units as the input layer:

%
\begin{equation}
\label{y} 
\mbox{y} = s(w_{y} h + b_{y}) 
\end{equation}
%
where $s(.)$ denotes the logistic sigmoid function $(1+\exp(-x))$. $w_{h}$
and $w_{y}$ represent the \textit{input to hidden} and the\textit{ hidden to output}
weights, respectively. In addition, $b_{h}$ and $b_{y}$ identify the
bias of the hidden and output units. With
the purpose of reconstructing the error
between $x$ and $y$, a metric based on the Euclidean distance is
generally minimized. This reconstruction loss is defined by:
%
\begin{equation}
\label{d} 
\mbox{d(x,y)} = {\Vert y-x \Vert}^{2} 
\end{equation}
%
A typical architecture of autoencoder is presented in Fig. \ref{ea}. A stacked Autoencoder (SAE) is a neural network built on the top of several layers of autoencoders where the output of each hidden layer is connected to the input of the next hidden layer. Fig. \ref{sea}
shows a simple representation of a SAE.

%
%

%
%

\subsubsection{CNNs}
\label{CNNs}

\begin{figure*}[!t]
	\centering
	\includegraphics[width=16cm]{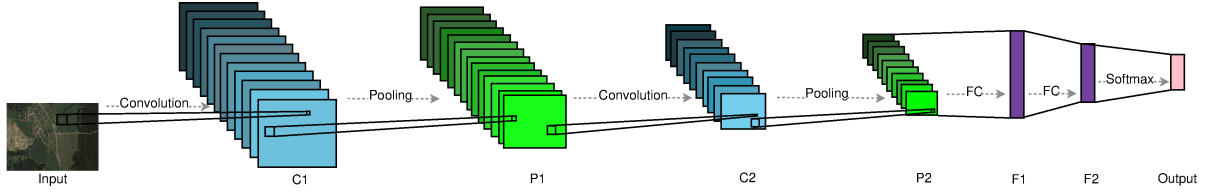}
	\caption{\label{cnn} A flowchart of a conventional CNN, which
		consists of two convolutional layers (C1, C2), two pooling layers
		(P1, P2), two fully connected layers (F1, F2) and a softmax layer
		(output).}
\end{figure*}

Convolutional neural networks, also known as CNNs
\cite{LeCunBoserDenkerEtAl89}, are a special form of neural network designed for  processing data that has a known grid-like representation,  for example image data, which can be considered  a two dimension (2D) grid of pixels. Generally,
the CNNs can be thought of as an extractor of hierarchical characteristics,
which, on the one hand, extracts features of diverse abstraction layers, and on the other hand, maps the raw pixel intensities into a feature vector. \cite{Zhang2016}. An architecture
of a typical  CNNs is illustrated in Fig. \ref{cnn}, where $conv$, $mp$, and
$fc$ denote convolutional, max-pooling and fully-connected layers,
respectively.
Convolutional layer represents the fundamental component of the CNN
architecture \cite{Yamashita2018}. In this layer, several trainable
convolution  kernels  (called also filters) are applied to the previous
layer. The weights of these kernels aim to connect units in a feature
map with the previous layer. As a result of convolution, local
conjunctions of features are detected and their appearance is mapped
to the feature maps. The stacking of various convolutional layers
increases the depth of networks which makes the extracted maps more
abstract. The earlier layers enhance features, for example edges, however,
the following layers aggregate these features in the form of motifs, parts, or
objects.   Formally, suppose that $m_{l}$ represents the filters convolution number in layer $l$ of the network, and $x^{n}_ {l-1}$ the 2D array
related to the $n-th$ input of layer $l$.  The $k-th$ output feature
vector  of layer $l$, denoted $z^{k}_{l}$, can be computed as follows:
%
\begin{equation}
\label{output_feature_map}
z^{k}_{l} = \Bigl [   \sum^{m_{l-1}}_{n=1} w^{k,n}_{l} \otimes x^{n}_{l-1}\Bigr ]  + b^{k}_{l}
\end{equation}
%
where $b^{k}_{l}$ is the bias matrix, $w^{ k,n}_{l}$ represents a filter
connecting the $n$-th feature map in the previous layer ($l-1$) with the
$k$-th feature map in layer $l$, and $\otimes$ denotes the convolution
operator.  Typically, after the convolution operation a nonlinear
activation function is performed on each element of the convolution
result.
%
\begin{equation}
\label{activation_function1}
y^{k}_{l} = f \Bigl (z^{k}_{l}   \Bigr )  
\end{equation}
%
A range of activation functions has been proposed in the literature
to improve the performance of CNNs, for example the sigmoid function
\cite{ITO1991385}, hyperbolic tangent function (tanh)
\cite{ANASTASSIOU20111111}, adaptive piecewise linear activation (APL)
\cite{Agostinelli2014LearningAF}, and the popular rectified linear
unit (ReLU) \cite{xu2015empirical}. The convolution process is
followed by a max-pooling operation. 
This step aims to replace the output of the network at some particular positions with a summary statistic relating to the neighborhood of  this location \cite{Goodfellow2016}. 
The pooling operation aims to gradually minimize, the spatial size of the output feature maps, and hence, decreases the parameters number of the network. Generally, there are two standard choices for the operation of pooling: max and average. Formally, for a $v \times  v$ window-size
neighbor represented by $N$. The average takes the arithmetic mean of
the elements in each pooling region as follows:
%
\begin{equation}
\label{activation_function2}
p^{k}_{j} = \vert \frac{1}{R^{k}_{j}} \vert \sum_{i\in R^{k}_{j} } y^{k}_{i,j}  
\end{equation}
%
while the max operation takes the largest element:
%
\begin{equation}
\label{activation_function3}
p^{k}_{j}  =  \max_{i \in R^{k}_{i}} y^{k}_{i,j}  
\end{equation}
%
where $R_j$ is pooling region (i.e., the number of elements in $N$)
and $y^{k}_{i,j}$ is the activation value related to the position $(i,
j)$.
After the pooling operation, the output feature maps of the previous layer
are flattened and provided to fully connected layers. These layers are exploited
to extract more  high-level information  by reshaping feature maps
into an $n$-dimension vector \cite{Li8697135}.
At the last layer of the network, called the classification layer,
neurons are gathered automatically into $C$ output feature maps that correspond to the
number of classes. 
Then, using a \textit{softmax}
function, the output of the classification layer $L$ is converted into (normalized) probability distribution errors. Specifically, the probability distribution of classes  is produced via the following function:
%
\begin{equation}
\label{probability_distribution}
p_{c}=\frac{\mbox{exp}(y^{c}_{L})}{\sum_{c^{'}=1}^{C}\mbox{exp}(y^{c^{'}}_{L})}
\end{equation}
%
where the calculated probabilities are within a $[0, 1]$ range, and
the sum of all the probabilities is equal to $1$.
Convolutional Neural Networks (CNNs) have been well established as a
powerful class of models from a variety of computer vision tasks
\cite{Andrej42455} including change detection in remote sensing
images. 
Hence, different successful CNNs architectures have been
suggested in the literature. The current surge of the CNNs in many
tasks heavily relies on the use of modern network architectures, such
as the AlexNet   \cite{NIPS2012Alex}, VGG  \cite{Simonyan15}, and
RESNET \cite{He7780459}. These modern architectures explore new and
innovative ways for constructing convolutional layers that guarantee
more efficient learning \cite{Jordan2018}.
%
%

\subsubsection{RNNs} 
\label{RNNs} 

\begin{figure}[!t]
	\centering
	\includegraphics[width=8.6cm, height=3cm]{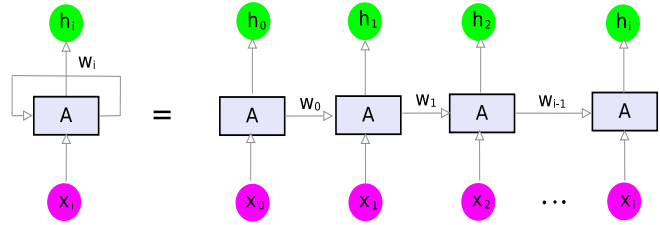}
	\caption{\label{rnn}An unrolled recurrent neural network.}
\end{figure}

Recurrent neural networks, also known as RNNs \cite{Rumelhart1988},
are a class of neural networks that allows processing sequential
data. Particularly, this model  is enhanced  by the integration of edges
that spanning adjacent time steps which introduces the notion of time
\cite{lipton2015}. Compared to the convolutional neural network that
is specialized for processing a grid of values $X$ such as an input
image, a recurrent neural network allows operating over a sequence  of
vectors or values with the help of a recurrent hidden state (see
Fig. \ref{rnn}). Formally, suppose that $x(1), . . . , x(n)$ is a sequence of
vectors where $x_{t}$ represents the data at the $t_{th}$ time
step. Two activation functions define all calculations required for
computation at each temporal sequence $t$:
%
\begin{equation}
\label{ht} 
\mbox h_{t} =f(W_{h}x_{t}+U_{h}h_{t-1}+b_{h})
\end{equation}
%
\begin{equation}
\label{yt} 
\mbox y_{t} = f2(W_{y}h_{t} + b_{y} )
\end{equation}
%
where $U_{h}$ is the same matrix utilized at each time step. Via this matrix,
the hidden units in the previous step  $h_{t-1}$ is used to compute
$h_{t}$, while the current observation provides a weighted term
$W_{h}x_{t}$, which is summed with $U_{h} h_{t-1}$ and a bias term
$b_{y}$. Both $W_{h}$ and $b_{y}$ are typically replicated over
time. The output layer is represented by a conventional neural network activation function applied to the linear transformation of the hidden units, and the process is repeated for every time phase \cite{WittenFrankHall11}.  Unfortunately, standard RNNs suffer from a
critical drawback related to the vanishing gradient problem,  which
makes the neural network hard to be trained properly. To overcome this
serious problem, long short-term memory (LSTM) \cite{HochSchm97} and
gated recurrent unit (GRU) \cite{chung2014empirical} were
suggested. 
One advantage of LSTM is that it introduces the notion of memory cell, a unit of computation that replaces classical nodes in the hidden layer of a network. This capability of memory cells  able to overcome difficulties with training encountered by earlier recurrent networks.
Like the LSTM unit, the GRU is characterized by units which control the information flow within the unit, nevertheless,  without having a distinct memory cell \cite{Junyoung2014}.

%
%

\subsubsection{GANs} 
\label{gans} 

\begin{figure}[!t]
	\centering
	\includegraphics[width=8.6cm, height=6cm]{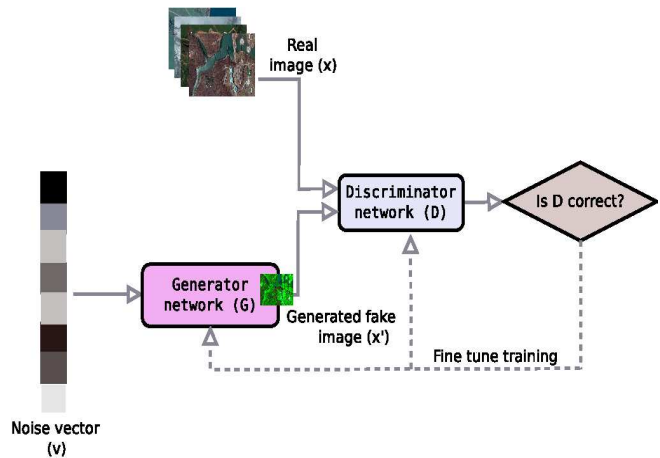}
	\caption{\label{gan} General illustration of the structure of a Generative Adversarial Network (GAN).}
\end{figure}

Generative adversarial networks (GANs) were proposed by
\textit{Goodfellow et al.} \cite{NIPS2014_Goodfellow}. Given a real
data (e.g., images), this generative technique learns to produce novel
data with the same statistics as the original data. GANs are based on
a game theoretical scenario in which the generator network must compete
against an adversary \cite{Goodfellow2016}. A general illustration of
the structure of a GAN is shown in Fig. \ref{gan}.  Formally, from
training data $x$ and a provided a priori distribution (i.e., random
noise) $v$, the generator network directly generates fake samples
$G(v)$. Its adversary, the discriminator network, aims to
differentiate between samples provided by the training data and
samples produced by the generator. While the discriminator $D$ is
trained to maximize the value of $\log(D(x))$, indicating the
probability of selecting the correct labels to the training samples, the
generator block $G$ is trained to minimize $\log(1-D(G(z))$  \cite{Li8697135}. Thus,
$D$ and $G$ play a two-player minimax game as follows:
%
\begin{equation}
\label{g-a} 
\mbox GAN = \arg \min_{D}  \max_{G} L_{GAN}(G, D)
\end{equation}
%
where
%
\begin{equation}
\label{Model}
\nonumber
\mbox L_{GAN}(G, D) \!\!\; = \!\;\;\mathbb{E}_{x \thicksim p(x)}[\log D(x)] 
\!\!\!  \; \; +  \mathbb{E}_{v \thicksim (v)} [\log (1 - D(G(v)))]
\end{equation}
%
Here, $\mathbb{E}$ denotes the expectation operator.  The main goal of
training generative networks is to produce examples that appear
realistic compared to the original data. Based on that assumption,
GANs have been successfully used in different computer vision and image
processing applications.
%

%
%
\section{Methods and data used to review DL for CD in remote sensing images}
\label{Methods_data}

\subsection{A meta-analysis process for data extraction}
\label{meta-analysis}

We searched and collected all published studies relevant to change
detection in remote-sensing images using the deep learning
approach. The search for studies was conducted using the web of
science database\footnotemark[2]. It is the most trusted publisher
global citation database. The generated dataset was built using an
advanced search option (search date: April $18$th, $2020$) with a
relevant controlled vocabulary included; deep learning, change
detection and remote sensing topic, etc. All of the studies included
in our research had been published up to $2014$. Ignored from the
search query were prefaces, article summaries, interviews,
discussions, news items, correspondences, readers' letters, comments, summaries of tutorials, panels, workshops, and
poster sessions. This search strategy resulted in a total of $160$
unique papers, including $110$ journal articles and $47$ conference
papers, two early access paper and one editorial material paper. All
these included studies are summarized in one file publicly accessible
via this link: "http://www.lazharkhelifi.com/?publications=rev\_cd\_dl.zip". It
is worth noting  that every article included in the review was read in
detail by the authors.  
%
\footnotetext[2]{The web of science database is accessible via 
	the following link: https://www.webofknowledge.com/
} 

\subsection{Referred journals and conference papers}
\label{journals}

Among the set of $110$ peer-reviewed journal papers, a larger part of
articles were published in the ten journals shown in Table
\ref{Tab1}. Note that journals with only one publication are not
listed here. Overall, these $10$ journals include $82$ articles
peer-reviewed journal papers related to DL change detection and remote
sensing. Regarding the number of articles published per journal, the
top five peer-reviewed journal papers are; \textit{Remote Sensing},
\textit{IEEE Transactions on Geoscience and Remote Sensing (TGRS)},
\textit{IEEE Access}, \textit{IEEE  Journal of Selected Topics in
	Applied Earth Observations and Remote Sensing} and \textit{IEEE
	Geoscience and Remote Sensing Letters}.

We found that the topic is now well represented at the major
international remote sensing conferences. Thus, among the set of  $47$
conference papers, a majority of the articles were published by the
two remote sensing academic societies listed in Table \ref{Tab2},
namely the \textit{IEEE Geoscience and Remote Sensing Society
	(IGARSS)} and \textit{the Society of Photo-Optical Instrumentation
	Engineers (SPIE)}. The conferences with only one publication are not
listed in this table. The reader should bear in mind that conference
papers were excluded from the present meta-analysis because many were
expanded into journal papers after presenting at the conferences (as
in \cite{MA2019166}). In addition, after a deep understanding of the
content of all the papers, $20$ journal papers that not cover the
subject of this study were also excluded.

\subsection{Brief interpretation of the results}
\label{Interpretation}

\begin{figure}[!t]
	\centering
	\includegraphics[width=8cm]{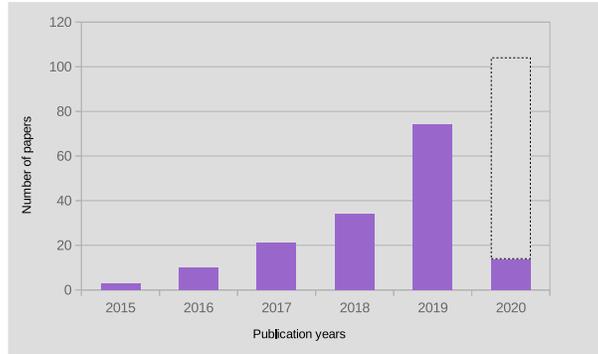}
	\caption{\label{stat} Growing number of published papers related to deep learning 
		for change detection in remote sensing (we predict more than 100 papers in 2020).}
\end{figure}

\begin{figure}[!t]
	\centering
	\includegraphics[width=8cm]{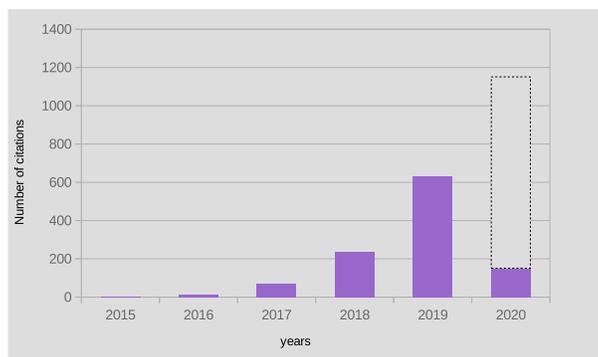}
	\caption{\label{citation} The number of  citations per year for papers related 
		to deep learning for change detection in remote sensing (we predict more than 1000 citations in 2020).}
\end{figure}

\begin{figure}[!t]
	\centering
	\includegraphics[width=8cm]{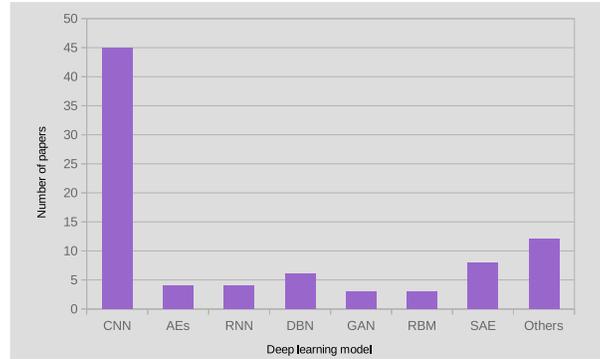}
	\caption{\label{model} Distribution of DL models used in the studies.}
\end{figure}

\begin{figure}[!t]
	\centering
	\includegraphics[width=8cm]{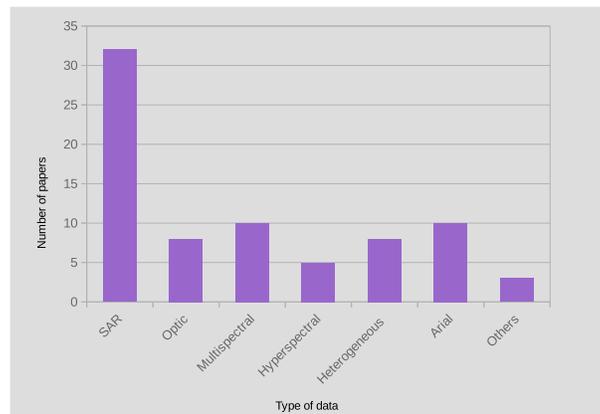}
	\caption{\label{type} Distribution of types of remote sensing images used in the studies.}
\end{figure}

Several general conclusions may be drawn from the conducted
statistical analysis to examine the trend in the use of DL for change
detection. Trends and projections are illustrated in this study using
histogram graphs in order to better visualize the distribution of the
data. Figure \ref{stat} reveals that there has been a marked increase
in the number of scientific papers released on the topic since $2015$. The
number of published papers is expected to grow even more tremendously
in the coming years.
Similarly, the graph of Fig. \ref{citation} shows that there has been
an important increase in the number of citations of those
papers. Table \ref{Tab3}  highlights the top three most-cited
papers. This exponential growth, both for the number of published
papers and the number of citations, validates the rapid growth of
interest in the study of deep learning for change detection in remote
sensing images. 
Notably, the number of journal papers on this topic now exceeds the
number of conference papers. This indicates the technical maturity of
this research area.
As can be seen from Fig. \ref{model}, the CNN model has been the most
widely applied for change detection, followed by the SAE, DBN, RNN,
AEs, RBM and GAN models. This higher popularity of CNN is probably
because it is more suitable to learn hierarchical image
representations from the input data by sequentially abstracting
higher-level features \cite{Yamashita2018}.
Looking at Fig. \ref{type}, it is apparent that the SAR image type has been the
most commonly used  within deep learning model for change detection,
followed by multispectral, arial, optic, heterogeneous (i.e.,
multi-modal), and hyperspectral images.  The reason for this is, that
synthetic-aperture radar captures images using microwave signals which
can enter through clouds \cite{Jaturapitpornchai_2019}, and is therefore more likely to have a significant
advantage of being insensitive to  sunlight and complex atmospheric conditions \cite{Yang20199}.

%
%
%
%
%
%
\begin{table*} [!t]
	\caption{\label{Tab1} Journals identified as pertinent, and number of relevant papers.}
	\renewcommand{\arraystretch}{1.4}
	\centering
	\begin{tabular}{|p{10cm}||c|}
		\hline\hline
		Name of journal  &  \# \\
		\hline
		-Remote Sensing- & (31)\\
		-IEEE Transactions on Geoscience and Remote Sensing (TGRS)-  & (13)\\
		-IEEE Access- & (7)\\
		-IEEE Journal of Selected Topics in Applied Earth Observations and mote Sensing-  & (7)\\
		-IEEE Geoscience and Remote Sensing Letters- & (5)\\
		-ISPRS Journal of Photogrammetry and Remote Sensing- &  (5)\\
		-Journal of Applied Remote Sensing- & (5)\\
		-IEEE Transactions on Neural Networks and Learning Systems- & (4)\\
		-Applied Sciences-Basel- & (3)\\
		-International Journal of Image and Data Fusion- & (2)\\
		\hline
		\hline
	\end{tabular}
\end{table*}

%
%
\begin{table*} [!t]
	\caption{\label{Tab2} Conferences and proceedings determined as pertinent, 
		and number of relevant papers.}
	\renewcommand{\arraystretch}{1.4}
	\centering
	\begin{tabular}{|p{10.7cm}||c|}
		\hline\hline
		Title of conference/Proceedings  &  \# \\
		\hline
		-International Geoscience and Remote Sensing Symposium (IGARSS)- & (17)\\
		-Proceedings of Society of Photo-Optical Instrumentation Engineers (SPIE)- & (9)\\
		\hline
		\hline
	\end{tabular}
\end{table*}

%
%
\begin{table*} [!t]
	\caption{\label{Tab3}
		The top three  most-cited papers.
	}
	\renewcommand{\arraystretch}{1.4}
	\centering
	\begin{tabular}{|p{2.6cm}||p{7cm}||p{3cm}||p{2cm}|}
		\hline\hline
		Authors &Title& Year of publication & Times cited 
		\\
		\hline
		-\textit{Gong et al.} \cite{7120131}-   
		& -Change Detection in Synthetic Aperture Radar Images Based on Deep Neural Networks- & 2016 & (224)\\
		-\textit{Lyu et al.} \cite{Lyu_2016}- 
		& -Learning a Transferable Change Rule from a Recurrent Neural Network for Land Cover Change Detection- & 2016 & (126)\\
		-\textit{Zhang et al.} \cite{ZHANG201624}- 
		& -Change detection based on deep feature representation and mapping transformation for multi-spatial-resolution 
		remote sensing images-  & 2016 & (108)\\
		\hline
		\hline
	\end{tabular}
\end{table*}

%
%
\section{Deep learning for change detection in remote Sensing images}
\label{Deep_learning}

Deep learning has recently become the focus of considerable interest in the change detection
field \cite{MA2019166}. It aims to automatically learn high-level features from various remote sensing data compared to traditional hand-crafted features-based methods \cite{Zhang2016}. 
Deep learning approaches for change detection can be
grouped in several ways by considering different perspectives. In this
study, therefore, the deep learning approaches used for change detection are
classified into three groups based on the learning technique and the
availability of a training data that can be either labeled or
unlabeled. The first type contains fully supervised methods which
solve the problem by learning from a labeled training dataset. The
second type of methods contains fully unsupervised methods that learn
from unlabeled datasets. Both supervised and unsupervised methods serve at selecting the available features that consistent with the target concept.
Hence, in supervised learning, the target concept is explicitly correlated to
class affiliation, while in unsupervised learning the target concept typically targeted  through  inherent structures of the data \cite{Zhao2007SpectralFS}. The third type of methods contains transfer
learning based methods. Transfer learning  is an important machine
learning technique which attempts to utilize the knowledge learned from one
task and to apply it on another, but associated, task 
with the purpose
to either reduce the necessary fine-tuning data size or improve
performances \cite{chung-etal-2018-supervised}. Sections
\ref{supervised}, \ref{unsupervised}, and \ref{Transfer} will outline
these types of methods in detail.

\subsection{Fully Supervised learning based-methods}
\label{supervised}

For a long time, it was commonly assumed that the process of training
deep supervised neural networks is challenging, time-consuming, and
too difficult to perform \cite{Zhang2016}. While the standard learning
strategy consisting of randomly initializing the weights, recently, it
was found that deep supervised networks can be trained by proper
weight initialization. This novel strategy  just adequate enough for improving the gradient flow as well as the transmission of useful information by the activations 
\cite{Bengio2013} \cite{7486259}.  The efficiency of supervised deep networks is particularly evident in case of the availability of large amount of labeled data used to properly train it.

In recent years, some pure supervised
DL methods have been suggested for change detection in RS images relying
on CNNs \cite{Penatti7301382} \cite{Zhang2016}. These CNNs based
studies have shown superior performances to the classical state-of-art
methods. The CNNs are hierarchical models which converts the input
image into multiple layers of feature maps.  These generated maps consist of high-level discriminatory features that reflect the original input data \cite{7486259}.  Based
on the fully convolutional networks U-Net is considered as one of the
standard CNNs architectures used for change detection task. The
general network architecture of U-Net is symmetric, having an
\textit{encoder} that extracts spatial features from the image, and a
\textit{decoder} that builds the segmentation map from the encoded
feature  \cite{Ronneberger2015}. 

\textit{Jaturapitpornchai et al.}  \cite{Jaturapitpornchai_2019} have
proposed in detail a U-Net-based network, which detects the novel
buildings construction in developing regions using two SAR images
captured at different times. Subsequently, the U-Net architecture was
extended through a few modifications in other works. In this regard,
\textit{Hamdi et al.}  \cite{Mahmoud2019} have developed an algorithm
using a modified U-Net model for automatic detection and
mapping of damaged areas in an ArcGIS environment. Their model was
trained based on a database of a forest area in Bavaria,
Germany. 
Recently, an improved UNet++ architecture was proposed by \textit{Peng et al.} 
\cite{Peng2019} for end-to-end change detection of VHR satellite images. 
In order, to learn multi-scale feature maps dense skip connections were 
established between the different layers of this  architecture. 
In addition, a residual block strategy was followed to facilitate 
gradient convergence of the  network.
For change detection in hyperspectral image, a general end-to-end two-dimensional
CNNs framework, called GETNET, was presented by \textit{Wang et al.} \cite{Wang2019}. 
In addition, a conventional change vector analysis (CVA) method \cite{Malila1980} was adopted to generate pseudo-training sets with labels.
\textit{Wiratama et al.} \cite{Wahyu2018} proposed a dual-dense
convolutional network for recognizing pixel-wise change on the basis of a
dissimilarity analysis of neighborhood pixels on high resolution
panchromatic (PAN) images. In their suggested algorithm, two fully
convolutional neural networks are utilized to compute the dissimilarity of
neighboring pixels. Further, a dense connection in convolution layers
is performed to reuse preceding feature maps by connecting them to all
subsequent layers.
\textit{Zhang et al.} \cite{ZhangChi2019} have introduced a fully
atrous convolutional neural network (FACNN). In this FACNN, first, an
encoder which consists of fully atrous convolution layers, is used for
extracting scale  features from VHR images. Afterwards, a change map based on pixel is generated using the classification map of
current images and an outdated land cover geographical information
system (GIS) map.
\textit{Daudt et al.} \cite{Daudt2019} have proposed an integrated
network based on deep FCNNs that performs a land cover mapping and
change detection simultaneously, using information from the land cover
mapping branches to help with change detection. 
\textit{Zhang et al.} \cite{Zhang20199} presented  a spectral-spatial
joint learning network (SSJLN). At the first part of this model, the spectral-spatial joint representation is derived from the network similar to the Siamese CNN (S-CNN) \cite{ZhenchaoZhang2018}. Second, these extracted features are combined together using a feature fusion block. To explore the underlying information of the combined features, discrimination learning is then performed at the last step.
\textit{Liu et al.} \cite{Liu2019} have demonstrated the
complementarity of CNNs and  bidirectional long short-term memory
network (BiLSTM) by combining them into one unified
architecture. While, the former is useful in extracting the rich spectral-spatial features from  bi-temporal images, the latter is powerful in analyzing the temporal dependence of bi-temporal images and transferring the features of images.  Similarly, \textit{Cao et
	al.} \cite{Cao2019a} have combined a deep denoising model trained on
a huge number of simulated SAR images patches with a CNNs
model. While, the deep denoising network is adopted to keep useful
information and suppress noise simultaneously, a three layers of a CNN
model are built to establish the feature learning process. 
Contrary
to previous approaches, that rely on CNNs based models
\textit{Wiratama et al.} \cite{Wahyu20199} have proposed a  fusion  architecture  combining front-end and back-end neural networks.  In order to accomplish low-level and high-level differential detection, the fusion network contains both single-path and dual-path networks. In addition , based on the two dual outputs, a
two-stage decision algorithm was proposed by authors to efficiently provide the final change detection result. This method has shown a good performance for the identification of changed/unchanged areas in high-resolution panchromatic images.

\subsection{Fully unsupervised learning based-methods}
\label{unsupervised}

Supervised deep learning methods such as the CNNs and its modified models
have achieved satisfactory result  in many computer vision tasks
due the availability  of large annotated datasets
\cite{Zhang2016}. Unfortunately, for change detection task, there are
often not enough training data to build such models. In addition, building
a ground-truth map reflecting the real change information of ground
objects costs lots of time and effort
\cite{Cao2019b}. Therefore, in many cases, it is more efficient to
learn the change features generated from a remote sensing image in an
unsupervised manner \cite{Lv2018}. 

Unsupervised feature-learning methods are mainly based on models which may learn feature representations from the patches (of images, par example) without any necessary supervision \cite{7486259}. There have been numerous enhancements and evolution to
the unsupervised deep learning approach that has been successfully
applied to recognizing remote sensing (RS) scenes and targets. One of
the most well-known and significant approaches is to stack (or to
combine) together different shallow feature-learning methods like the Gaussian Mixture model, AEs, sparse coding and RBMs
\cite{Zhang2016}.
In this regard, for change detection in  multispectral images, \textit{Zhang et al.} \cite{Zhangh2016} have proposed a new unsupervised method
combining the DBN and the feature change analysis (FCA).  Thus, to capture the useful information for discrimination between changed and unchanged regions and  to also suppress the irrelevant variations, the available spectral channels are transformed into an abstract feature space via the DBN.  Then, using these learned  features, an FCA is performed to identify the different types of change. 
Similarly, \textit{Su et al.} \cite{Su2017} have introduced a novel
deep learning and mapping (DLM) framework oriented to the ternary change detection task for information unbalanced images. In their method, two  types of neural networks are used. First, a  stacked denoising autoencoder is applied to two input  images, serving as a feature extractor.  Then, after a selection  step of relevant samples,  mapping functions are generated  by a stacked mapping network, establishing the relationship between the features of each class. 
Afterwards, a comparison between the features is performed and the final ternary map is generated via a clustering process of the comparison result.
\textit{Gao et al.} \cite{Feng2017} have proposed a novel SAR image
change detection method based on deep semi-nonnegative matrix
factorization (Deep Semi-NMF) \cite{Wang6165290} and singular value
decomposition (SVD) networks \cite{XueLG13}. In their suggested method, the deep Semi-NMF is used as a pre-classification step. Following this, the SVD network of two SVD convolutional layers is applied to obtain reliable features, where good quality of these obtained features effectively improves the classification performance.
To achieve more precise ternary change detection without any supervision, \textit{Gong et al.}\cite{GONG2017212} have combined SAE, CNN and an unsupervised clustering algorithm.  First,  noise  is removed and  key change information are extracted by transforming difference image into a suitable feature space using SAE.
Next, an unsupervised clustering is established  on the feature maps learned by SAE. This final step aims to provide reliable pseudo labels for training the CNN as a change feature classifier. 
\textit{Lv et al.} \cite{Lv2018} have presented a feature learning
method based on the combination of a stacked contractive autoencoder
(sCAE) and a simple clustering algorithm. In this method, first, an affiliated temporal change image is built using three different metrics. the aim of this strategy is to provide more information about the temporal difference on the pixel level.
Second, homogeneous change samples are provided by generating a set of superpixels using a  simple linear iterative clustering algorithm. Third, these  generated superpixel-samples  are used as input to train  a sCAE network.  Then, the encoded features results from the sCAE model are binary classified to create the change result map.
\textit{Gong et al.} \cite{Gongg2019} have developed  a generative
discriminatory classified network (GDCN) for multispectral image change detection. The generative adversarial networks represent the key block of this proposed model by  providing three types of data; labeled data, unlabeled data, and new fake data. More precisely, this GDCN composes of a discriminatory classified network (DCN) and a generator (G).  While the DCN divides the input data into changed class, unchanged class, and extra class (i.e., fake class), the generator recovers the real data from input noises to provide additional training samples. Finally, the bitemporal multispectral images are fed to the DCN to get a final reliable change map. 
For  change detection in SAR images, \textit{Gen et al.} \cite{Geng2019} have proposed SGDNNs, an unsupervised saliency guided deep neural networks. The first step in this model consists of extracting a salient region from the difference image (DI), which probably belongs to the changed object. Then,  a hierarchical fuzzy C-means (HFCM) clustering \cite{Geva19999} is established to select samples with higher probabilities to be changed and unchanged. Using these pseudotraining samples, a DNNs based on the nonnegative-and Fisher-constrained autoencoder are applied to get reliable final detection.
\textit{Li et al.} and \cite{Xuelong2019} performed  change detection for
hyperspectral images using a novel noise modeling-based unsupervised
fully convolutional network (FCN) framework. Specifically, their suggested deep CNN is trained  using the change detection maps of existing unsupervised change detection methods, while the noise is removed during the end-to-end training process. 
Recently, \textit{Huang et al.} \cite{Huang2019} have proposed a new
unsupervised algorithm based on deep learning called ABCDHIDL to
automatically detect the building changes from multi-temporal
high-resolution remote sensing (HRRS) images. In this algorithm,
initially, a convolution operation is adopted for two reasons;  first,
to extract the spatial, texture and spectral features and second to
generate a combined low-level feature vector for each pixel. Then,
the unlabeled samples are injected to pre-train a DBN network, where
its parameters are optimized by jointly using the extreme learning
machine (ELM) classifier \cite{SURESH2009541}. To further improve the
detection process, labeled samples are offered by an automatic
selection based on a morphological operation.

\subsection{Deep transfer learning based-methods}
\label{Transfer}

In many remote sensing applications, it is so  expensive or impossible to
recollect the required training data and rebuild the models
\cite{Pan2010}. In particular, for the change detection task, there
are often not enough training data that accurately represent the real change
information of ground objects. 
Therefore, it is important to reduce
the requirement and effort to recollect the training data. In that
context, transfer learning or knowledge transfer among task domains
can be a reliable solution. 

Transfer learning is defined as the
capability of extracting knowledge from one or more source tasks and
applying it to a novel or target task \cite{Pan2010}.
Formally, given a source domain $D_S$ with a related source task $T_S$
and a target domain $D_T$ with a corresponding task $T_T$, transfer
learning is the proceeding of improving the target predictive function
$f_T(.)$ by utilizing the corresponding information from $D_S$ and
$T_S$, where $D_S$ $\neq$ $D_T$ or $T_S$ $\neq$ $T_T$
\cite{Weiss2016}. 

There are two basic approaches currently being adopted in research
into transfer learning. The first approach consists of using the
outputs of one or more layers of a network (such as AlexNet or
resnet-101) trained on a different task as generic high dimensional
feature detectors and training a new shallow model based on these
features \cite{zhao2018object}.  The second approach is more involved,
which consists of fine-tuning the network pre-trained in general
images. Hence, final layer (for classification/regression) is not just
replaced, but also,  previous layers are retained again
\cite{ABDALLA2019105091}. 
Following this former approach, \textit{Hou et al.} \cite{Bin2017}
have transferred a CNNs already pre-trained on large-scale natural
image data set (e.g., ImageNet \cite{imagenet_cvpr09}), to a RS
domain. Specifically, to get better results they fine-tune the VGG-16
\cite{Simonyan15} to adapt it to their optical RS images on an aerial
image dataset (AID) \cite{7907303}. 
Similarly, \textit{Venugopal et al.} \cite{Venugopal2019}, have
resorted to a  ResNet-$101$ \cite{Kaiming2015} network as a pretrained
model, and they fine-tuned parameters based on a dilated convolutional
neural network (DCNN) which detects the changes between the two
images. Afterwards, the classified result is determined from the final
feature map as unchanged and changed areas.
To solve the change detection problem  in optical aerial images, \textit{Zhang et al.} \cite{ZhangM2019} proposed a new method based on deep Siamese semantic network trained using an  improved triplet loss function.  First,  a DeepLabv2 \cite{Liang} model  pretrained on large-scale image data set (e.g., PASCAL VOC 2012 dataset \cite{pascal-voc-2012}), was transferred to the network, due to the  difficulty of directly training  the Siamese network. Based on this strategy, the network has achieved a comparable performance with limited computational cost and minimum training samples.
This change detection method is based on four steps; First, In order to perform a radiant correction to the two coregistered images,  the input bitemporal images are preprocessed using histogram matching. Second, the
preprocessed pair images are fed to the deep Siamese semantic network
in order to generate two feature maps. Following this, a resizing
operation is  applied for two semantic feature maps by a bilinear
interpolation. Afterwards, a distance map is obtained by computing the
Euclidean distance between semantic feature maps.  Finally, a simple threshold segmentation method is used to  separate the distance map, and therefore, to generate the final change detection result.
\textit{Fang et al.} \cite{Fang2019DualLS} proposed a novel hybrid
end-to-end framework named dual learning-based Siamese framework
(DLSF) for change detection from very high resolution (VHR)
images. This framework consists of two parallel streams which are dual
learning-based domain transfer and Siamese-based change
decision. While the first path is aimed at reducing the domain differences between two paired images and maintaining the intrinsic information by translating them into each other's domain, the second path is aimed at learning a decision strategy to decide the changes in two domains, respectively.
\textit{Yang, et al.} \cite{Yang20199} have adopted the concept of
change that is learned from the source domain to the target domain by
reducing the distribution discrepancy between two domains. In their
model, the pretraining stage includes two tasks; a supervised change
detection in the source domain using U-Net architecture and a
reconstruction network in the target domain without labels. The lower layers are shared between the two tasks, however,  the final layers related to each task are trained separately. After the pretraining step, reliable labels that are chosen from a CD map, are used to fine-tune the change detection network for the target domain.  
Although training data are limited in the task of sea ice change
detection, in the work of  \textit{Gao et al. } \cite{Gao2019} a large
data set was used to train a transferred multilevel fusion network
MLFN, in addition, a fine-tune strategy was utilized to optimize the
network parameters.
%

%
%
\section{Promising research directions}
\label{Promising_research}

To  advance the progress of the change detection task, in this section, we suggest two important directions for research, specifically
deep reinforcement learning and weakly supervised change detection.

\subsection{Deep reinforcement learning}
\label{reinforcement}

Due to the lack of sufficient labeled training databases for the supervised
change detection task, the description capability of the features
generated by deep learning methods may become limited or even
impoverished. Recently, deep reinforcement learning \cite{Mnih2015}
\cite{7989381} \cite{9043893} has become the focus of considerable interest in the field of machine learning and
has shown an excellent
potential and great performance  in various domains of computer vision
such as autonomous driving \cite{8911507} \cite{8998330}, object
tracking \cite{TENG2020107188} \cite{8642452}, person
re-identification \cite{Liu_2019_ICCV} \cite{8666162}, etc.

Deep reinforcement learning combines deep neural networks with a
reinforcement learning architecture, where intelligent machines can
learn from their actions similar to the way humans learn from
experience. Reinforcement learning enables software-defined agents to
learn from the environment on the basis
of random  exploration and to adjust the best possible actions based on
continuous feedback in order to attain their goals. Actions that get
them to the target outcome are rewarded (i.e., exploitation)
\cite{8694781}.
Formally, it consists of a finite number of states $s_i$ which
represent agents and the environment, actions $a_i$ realized 
by the
agent, probability $P_a$ of moving from one state to another on the basis of
action $a_i$, and reward $R_a(s_i,s_{i+1})$ corresponded to the move
to the next state $s_{i+1}$ with action $a$. 
To predict the best action as given by the function $D(s, a)$, balancing and maximizing the current reward $R$ and future reward $\alpha$ $ \cdot$ $\max[D (s^\prime , a^\prime ]$ is necessary.  Where $\alpha$  in the equation
denotes a fixed discount factor. Hence, this function $D (s, a)$ is represented as the
summation of current reward $R$ and future reward $\alpha$  $ \cdot $
$\max[D( s^\prime , a^\prime ) ]$ in the following way \cite{8694781}:

%
\begin{equation}
\label{g-b} 
\mbox D (s, a) = R + \alpha \cdot \max [D (s^\prime , a^\prime) ]
\end{equation}
%
%

Reinforcement learning is  particularly dedicated to solve problems consisting of both short-term and long-term rewards, for example, games such as go and  chess, etc.  However, combining reinforcement learning  and deep network architecture together yields deep reinforcement learning (DRL), which extends the use of reinforcement learning to robustly solve more difficult games and other challenging problems \cite{Volodymyr2016}.
Deep reinforcement learning not only  provides rich representations
characterized by a higher number of hidden layers of deep networks,  but also, presents a reinforcement learning-based Q-learning algorithm \footnotemark[3] that maximizes the reward for actions taken by the agent \cite{Volodymyr2016}. \textit{Fu et al.} \cite{Fu2018ASR}  have shown
the feasibility of using deep reinforcement learning for remote
sensing ship detection task. Recently, \textit{Li et al.}
\cite{Li2018AnAD} have proposed an interesting aircraft detection
framework based on the combination of a CNN model with reinforcement learning. Similarly, the change detection process can be solved as an
action-decision problem based on a sequence of actions refining the
size of the changed regions between two input images.
%
\footnotetext[3]{Q-learning is a reinforcement learning algorithm  required to find an optimal action-selection strategy to maximize the sum of the discounted rewards. \cite{PENG1994226}.}

\subsection{Weakly supervised change detection}
\label{Weakly}

Considering the high cost of the data labeling operation,  in many computer vision tasks, it is hard to get strong supervision information, (e.g., a dataset with fully ground-truth labels) \cite{Zhou2017}. 
Notably, in remote sensing images, the manual
annotation of objects is generally expensive and sometimes
unreliable. Particularly for the change detection task, the changed
regions are very small, the background is often cluttered and complex,
and the images may be taken by different sensors
\cite{CHENG201611}. However, training a change detection framework
based on weakly supervised learning (WSL) can alleviate the need for
manual annotation. Weakly supervised data include a small quantity of accurate label information, that differs from data in traditional supervised learning \cite{Li8735810}. In general, there are
three classes of weak supervision \cite{Zhou2017}:
%
\begin{itemize} 
	\item
	Incomplete supervision when a minimum quantity  data (among the training data) is provided with labels, which is inadequate to successfully train a learner.
	\item 
	Inexact supervision is when some supervision information is available, however, not as accurate as required (i.e., only coarse-grained label information is provided).
	\item 
	Inaccurate supervision relates the case in which the outlined labels are not really ground-truth and suffer from errors (i.e., learning with label noise).

\end{itemize}
%
Recent progress in the geospatial object detection field
\cite{Zhou2016WeaklyST} \cite{7153907} has shown the feasibility of
using weakly supervised learning. Similarly, 
it will be interesting to explore the potential of WSL-based change detection models  accurately for identifying the changed regions between two images. However, the
performance of existing WSL-based methods in remote sensing images is
still far from satisfactory. For example, accurate position of the
change cannot be yielded in detection of building changes
\cite{Jiang20119}.  Much effort also needs to be made to establish more efficient methods to improve the detection accuracy \cite{CHENG201611}.

%
%
\section{Conclusion} 
\label{Conclusion}

Recently, deep learning-based change detection in remote sensing field
has drawn significant attention and obtained good
performances. Deep learning based methods can automatically learn
complex features of remote sensing images on the basis of a huge number of
hierarchical layers, in contrast to traditional hand-crafted feature-based 
methods. In this work, publications related to DL in
remote sensing images were systematically analyzed through a
metaanalysis. In addition, a deeper review was conducted to describe
and discuss the use of DL algorithms specifically in the field of
change detection, which differentiates our study from previous reviews
on DL and remote sensing. Thus, several deep models that are often
used for change detection are described. In addition, we concentrate  on deep learning-based change
detection approaches for remote sensing images by providing a
general overview of the existing methods. Specifically, these deep
learning-based methods were divided into three groups; fully
supervised learning-based methods, fully unsupervised learning-based
methods and transfer learning-based methods. Besides, we have also
proposed two promising  future research directions.
Therefore, a further study with more focus on deep reinforcement learning and weakly supervised change detection methods  are strongly suggested.



\begin{thebibliography}{100}
	\providecommand{\url}[1]{#1}
	\csname url@samestyle\endcsname
	\providecommand{\newblock}{\relax}
	\providecommand{\bibinfo}[2]{#2}
	\providecommand{\BIBentrySTDinterwordspacing}{\spaceskip=0pt\relax}
	\providecommand{\BIBentryALTinterwordstretchfactor}{4}
	\providecommand{\BIBentryALTinterwordspacing}{\spaceskip=\fontdimen2\font plus
		\BIBentryALTinterwordstretchfactor\fontdimen3\font minus
		\fontdimen4\font\relax}
	\providecommand{\BIBforeignlanguage}[2]{{%
			\expandafter\ifx\csname l@#1\endcsname\relax
			\typeout{** WARNING: IEEEtran.bst: No hyphenation pattern has been}%
			\typeout{** loaded for the language `#1'. Using the pattern for}%
			\typeout{** the default language instead.}%
			\else
			\language=\csname l@#1\endcsname
			\fi
			#2}}
	\providecommand{\BIBdecl}{\relax}
	\BIBdecl
	
	\bibitem{LeCun2015}
	L.~Yann, B.~Yoshua, and H.~Geoffrey, ``Deep learning,'' \emph{Nature}, vol.
	521, pp. 436 -- 444, 2015.
	
	\bibitem{rish2001empirical}
	I.~Rish, ``An empirical study of the naive bayes classifier,'' in \emph{IJCAI
		2001 workshop on empirical methods in artificial intelligence}, vol.~3,
	no.~22.\hskip 1em plus 0.5em minus 0.4em\relax IBM New York, 2001, pp.
	41--46.
	
	\bibitem{Lewis1998}
	D.~D. Lewis, ``Naive (bayes) at forty: The independence assumption in
	information retrieval,'' in \emph{Machine Learning: ECML-98}, C.~N{\'e}dellec
	and C.~Rouveirol, Eds.\hskip 1em plus 0.5em minus 0.4em\relax Berlin,
	Heidelberg: Springer Berlin Heidelberg, 1998, pp. 4--15.
	
	\bibitem{Suykens99leastsquares}
	J.~Suykens, L.~Lukas, P.~V. Dooren, B.~D. Moor, and J.~Vandewalle, ``Least
	squares support vector machine classifiers: a large scale algorithm,'' 1999.
	
	\bibitem{CauwenberghsP00}
	G.~Cauwenberghs and T.~A. Poggio, ``Incremental and decremental support vector
	machine learning,'' in \emph{NIPS}, T.~K. Leen, T.~G. Dietterich, and
	V.~Tresp, Eds.\hskip 1em plus 0.5em minus 0.4em\relax MIT Press, 2000, pp.
	409--415.
	
	\bibitem{breiman2001random}
	L.~Breiman, ``Random forests,'' \emph{Machine Learning}, vol.~45, no.~1, pp.
	5--32, 2001.
	
	\bibitem{GISLASON2006294}
	P.~O. Gislason, J.~A. Benediktsson, and J.~R. Sveinsson, ``Random forests for
	land cover classification,'' \emph{Pattern Recognition Letters}, vol.~27,
	no.~4, pp. 294 -- 300, 2006.
	
	\bibitem{Safavian1991}
	S.~R. {Safavian} and D.~{Landgrebe}, ``A survey of decision tree classifier
	methodology,'' \emph{IEEE Transactions on Systems, Man, and Cybernetics},
	vol.~21, no.~3, pp. 660--674, 1991.
	
	\bibitem{FRIEDL1997}
	M.~Friedl and C.~Brodley, ``Decision tree classification of land cover from
	remotely sensed data,'' \emph{Remote Sensing of Environment}, vol.~61, no.~3,
	pp. 399 -- 409, 1997.
	
	\bibitem{Voulodimos}
	A.~Voulodimos, N.~Doulamis, A.~Doulamis, and E.~Protopapadakis, ``Deep learning
	for computer vision: A brief review,'' \emph{Computational Intelligence and
		Neuroscience}, vol. 2018, pp. 1--13, 02 2018.
	
	\bibitem{6639344}
	L.~{Deng}, G.~{Hinton}, and B.~{Kingsbury}, ``New types of deep neural network
	learning for speech recognition and related applications: an overview,'' in
	\emph{2013 IEEE International Conference on Acoustics, Speech and Signal
		Processing}, 2013, pp. 8599--8603.
	
	\bibitem{Palangi2016}
	H.~{Palangi}, L.~{Deng}, Y.~{Shen}, J.~{Gao}, X.~{He}, J.~{Chen}, X.~{Song},
	and R.~{Ward}, ``Deep sentence embedding using long short-term memory
	networks: Analysis and application to information retrieval,'' \emph{IEEE/ACM
		Transactions on Audio, Speech, and Language Processing}, vol.~24, no.~4, pp.
	694--707, 2016.
	
	\bibitem{SINGH1989}
	A.~SINGH, ``Review article digital change detection techniques using
	remotely-sensed data,'' \emph{International Journal of Remote Sensing},
	vol.~10, no.~6, pp. 989--1003, 1989.
	
	\bibitem{GONG2017212}
	M.~Gong, H.~Yang, and P.~Zhang, ``Feature learning and change feature
	classification based on deep learning for ternary change detection in sar
	images,'' \emph{ISPRS Journal of Photogrammetry and Remote Sensing}, vol.
	129, pp. 212 -- 225, 2017.
	
	\bibitem{9037317}
	R.~{Liu}, D.~{Jiang}, L.~{Zhang}, and Z.~{Zhang}, ``Deep depthwise separable
	convolutional network for change detection in optical aerial images,''
	\emph{IEEE Journal of Selected Topics in Applied Earth Observations and
		Remote Sensing}, vol.~13, pp. 1109--1118, 2020.
	
	\bibitem{Bovolo2007}
	F.~{Bovolo} and L.~{Bruzzone}, ``A split-based approach to unsupervised change
	detection in large-size multitemporal images: Application to tsunami-damage
	assessment,'' \emph{IEEE Transactions on Geoscience and Remote Sensing},
	vol.~45, no.~6, pp. 1658--1670, 2007.
	
	\bibitem{Coppin2004}
	P.~Coppin, I.~Jonckheere, K.~Nackaerts, B.~Muys, and E.~Lambin, ``Digital
	change detection methods in ecosystem monitoring: a review,''
	\emph{International Journal of Remote Sensing}, vol.~25, no.~9, pp.
	1565--1596, 2004.
	
	\bibitem{FERANEC2007234}
	J.~Feranec, G.~Hazeu, S.~Christensen, and G.~Jaffrain, ``Corine land cover
	change detection in europe (case studies of the netherlands and slovakia),''
	\emph{Land Use Policy}, vol.~24, no.~1, pp. 234 -- 247, 2007.
	
	\bibitem{VIANA2019621}
	C.~M. Viana, S.~Oliveira, S.~C. Oliveira, and J.~Rocha, ``29 - land use/land
	cover change detection and urban sprawl analysis,'' in \emph{Spatial Modeling
		in GIS and R for Earth and Environmental Sciences}, H.~R. Pourghasemi and
	C.~Gokceoglu, Eds.\hskip 1em plus 0.5em minus 0.4em\relax Elsevier, 2019, pp.
	621 -- 651.
	
	\bibitem{MA2019166}
	``Deep learning in remote sensing applications: A meta-analysis and review,''
	\emph{ISPRS Journal of Photogrammetry and Remote Sensing}, vol. 152, pp. 166
	-- 177, 2019.
	
	\bibitem{8113128}
	X.~X. {Zhu}, D.~{Tuia}, L.~{Mou}, G.~{Xia}, L.~{Zhang}, F.~{Xu}, and
	F.~{Fraundorfer}, ``Deep learning in remote sensing: A comprehensive review
	and list of resources,'' \emph{IEEE Geoscience and Remote Sensing Magazine},
	vol.~5, no.~4, pp. 8--36, 2017.
	
	\bibitem{7486259}
	L.~{Zhang}, L.~{Zhang}, and B.~{Du}, ``Deep learning for remote sensing data: A
	technical tutorial on the state of the art,'' \emph{IEEE Geoscience and
		Remote Sensing Magazine}, vol.~4, no.~2, pp. 22--40, 2016.
	
	\bibitem{Liu2015}
	S.~{Liu}, L.~{Bruzzone}, F.~{Bovolo}, and P.~{Du}, ``Hierarchical unsupervised
	change detection in multitemporal hyperspectral images,'' \emph{IEEE
		Transactions on Geoscience and Remote Sensing}, vol.~53, no.~1, pp. 244--260,
	2015.
	
	\bibitem{Yang2019_bb}
	G.~{Yang}, H.~{Li}, W.~{Wang}, W.~{Yang}, and W.~J. {Emery}, ``Unsupervised
	change detection based on a unified framework for weighted collaborative
	representation with rddl and fuzzy clustering,'' \emph{IEEE Transactions on
		Geoscience and Remote Sensing}, vol.~57, no.~11, pp. 8890--8903, 2019.
	
	\bibitem{Liu2019}
	R.~{Liu}, Z.~{Cheng}, L.~{Zhang}, and J.~{Li}, ``Remote sensing image change
	detection based on information transmission and attention mechanism,''
	\emph{IEEE Access}, vol.~7, pp. 156\,349--156\,359, 2019.
	
	\bibitem{Kevin2018}
	K.~L. de~Jong and A.~S. Bosman, ``Unsupervised change detection in satellite
	images using convolutional neural networks,'' \emph{CoRR}, vol.
	abs/1812.05815, 2018.
	
	\bibitem{Kadhim2016}
	N.~Kadhim, M.~Mourshed, and M.~Bray, ``Advances in remote sensing applications
	for urban sustainability,'' \emph{Euro-Mediterranean Journal for
		Environmental Integration}, vol.~1, no.~7, 2016.
	
	\bibitem{ZhangDianjun2016}
	Z.~Dianjun and Z.~Guoqing, ``Estimation of soil moisture from optical and
	thermal remote sensing: A review,'' \emph{Sensors (Basel)}, vol.~16, no.~8,
	2016.
	
	\bibitem{Sublime_2019}
	J.~Sublime and E.~Kalinicheva, ``Automatic post-disaster damage mapping using
	deep-learning techniques for change detection: Case study of the tohoku
	tsunami,'' \emph{Remote Sensing}, vol.~11, no.~9, p. 1123, May 2019.
	
	\bibitem{Kolos2019}
	M.~Kolos, A.~Marin, A.~Artemov, and E.~Burnaev, ``Procedural synthesis of
	remote sensing images for robust change detection with neural networks,'' in
	\emph{Advances in Neural Networks -- ISNN 2019}, H.~Lu, H.~Tang, and Z.~Wang,
	Eds.\hskip 1em plus 0.5em minus 0.4em\relax Cham: Springer International
	Publishing, 2019, pp. 371--387.
	
	\bibitem{Goodfellow2016}
	I.~Goodfellow, Y.~Bengio, and A.~Courville, \emph{Deep Learning}.\hskip 1em
	plus 0.5em minus 0.4em\relax The MIT Press, 2016.
	
	\bibitem{reed99a}
	R.~D. Reed and R.~J. Marks, \emph{Neural Smithing: Supervised Learning in
		Feedforward Artificial Neural Networks}.\hskip 1em plus 0.5em minus
	0.4em\relax {MIT} Press, 1999.
	
	\bibitem{McCullochPitts43}
	W.~S. McCulloch and W.~Pitts, ``A logical calculus of the ideas immanent in
	nervous activity,'' \emph{Bulletin of Mathematical Biophysics}, vol.~5, pp.
	115--133, 1943.
	
	\bibitem{hebb:behavior}
	D.~O. Hebb, \emph{The Organization of Behavior: {A} Neuropsychological
		Theory}.\hskip 1em plus 0.5em minus 0.4em\relax New York: Wiley, 1949.
	
	\bibitem{rosenblatt1958perceptron}
	F.~Rosenblatt, ``{The perceptron: A probabilistic model for information storage
		and organization in the brain.}'' \emph{Psychological Review}, vol.~65,
	no.~6, pp. 386--408, 1958.
	
	\bibitem{Rumelhart:1986we}
	D.~E. Rumelhart, G.~E. Hinton, and R.~J. Williams, ``{Learning Representations
		by Back-propagating Errors},'' \emph{Nature}, vol. 323, no. 6088, pp.
	533--536, 1986.
	
	\bibitem{Rumelhart1988}
	D.~E.~Rumelhart, G.~E.~Hinton, and R.~J.~Williams, ``Neurocomputing:
	Foundations of research,'' J.~A. Anderson and E.~Rosenfeld, Eds.\hskip 1em
	plus 0.5em minus 0.4em\relax Cambridge, MA, USA: MIT Press, 1988, ch.
	Learning Representations by Back-propagating Errors, pp. 696--699.
	
	\bibitem{Hinton:2006}
	G.~E. Hinton and R.~R. Salakhutdinov, ``Reducing the dimensionality of data
	with neural networks,'' \emph{Science}, vol. 313, pp. 504 -- 507, 2006.
	
	\bibitem{Bengio:LeCun:07}
	Y.~Bengio and Y.~LeCun, ``Scaling learning algorithms towards ai,'' in
	\emph{Large Scale Kernel Machines}, L.~Bottou, O.~Chapelle, D.~DeCoste, and
	J.~Weston, Eds.\hskip 1em plus 0.5em minus 0.4em\relax Cambridge, MA: MIT
	Press, 2007.
	
	\bibitem{Mohamed5704567}
	A.~{Mohamed}, G.~E. {Dahl}, and G.~{Hinton}, ``Acoustic modeling using deep
	belief networks,'' \emph{IEEE Transactions on Audio, Speech, and Language
		Processing}, vol.~20, no.~1, pp. 14--22, 2012.
	
	\bibitem{Mohamed5947494}
	A.~{Mohamed}, T.~N. {Sainath}, G.~{Dahl}, B.~{Ramabhadran}, G.~E. {Hinton}, and
	M.~A. {Picheny}, ``Deep belief networks using discriminative features for
	phone recognition,'' in \emph{2011 IEEE International Conference on
		Acoustics, Speech and Signal Processing (ICASSP)}, 2011, pp. 5060--5063.
	
	\bibitem{Li8697135}
	S.~{Li}, W.~{Song}, L.~{Fang}, Y.~{Chen}, P.~{Ghamisi}, and J.~A.
	{Benediktsson}, ``Deep learning for hyperspectral image classification: An
	overview,'' \emph{IEEE Transactions on Geoscience and Remote Sensing},
	vol.~57, no.~9, pp. 6690--6709, 2019.
	
	\bibitem{Buckner}
	\BIBentryALTinterwordspacing
	C.~{Buckner} and J.~{Garson}, ``Connectionism,'' \emph{E. N. Zalta (Ed.), The
		stanford encyclopedia of philosophy (Fall 2019). Stanford: Metaphysics
		Research Lab, Stanford University.}, 2019. [Online]. Available:
	\url{https://plato.stanford.edu/entries/connectionism/}
	\BIBentrySTDinterwordspacing
	
	\bibitem{ZABALZA20161}
	J.~Zabalza, J.~Ren, J.~Zheng, H.~Zhao, C.~Qing, Z.~Yang, P.~Du, and
	S.~Marshall, ``Novel segmented stacked autoencoder for effective
	dimensionality reduction and feature extraction in hyperspectral imaging,''
	\emph{Neurocomputing}, vol. 185, pp. 1 -- 10, 2016.
	
	\bibitem{LeCunBoserDenkerEtAl89}
	Y.~LeCun, B.~Boser, J.~S. Denker, D.~Henderson, R.~E. Howard, W.~Hubbard, and
	L.~D. Jackel, ``Backpropagation applied to handwritten zip code
	recognition,'' \emph{Neural Computation}, vol.~1, pp. 541--551, 1989.
	
	\bibitem{Zhang2016}
	L.~{Zhang}, F.~{Yang}, Y.~{Daniel Zhang}, and Y.~J. {Zhu}, ``Road crack
	detection using deep convolutional neural network,'' in \emph{2016 IEEE
		International Conference on Image Processing (ICIP)}, Sep. 2016, pp.
	3708--3712.
	
	\bibitem{Yamashita2018}
	Y.~Rikiya, D.~R.~K. Gian, and T.~Kaori, ``Convolutional neural networks: an
	overview and application in radiology,'' \emph{Insights into Imaging},
	vol.~9, pp. 115--133, 2018.
	
	\bibitem{ITO1991385}
	Y.~Ito, ``Representation of functions by superpositions of a step or sigmoid
	function and their applications to neural network theory,'' \emph{Neural
		Networks}, vol.~4, no.~3, pp. 385 -- 394, 1991.
	
	\bibitem{ANASTASSIOU20111111}
	G.~A. Anastassiou, ``Univariate hyperbolic tangent neural network
	approximation,'' \emph{Mathematical and Computer Modelling}, vol.~53, no.~5,
	pp. 1111 -- 1132, 2011.
	
	\bibitem{Agostinelli2014LearningAF}
	F.~Agostinelli, M.~D. Hoffman, P.~J. Sadowski, and P.~Baldi, ``Learning
	activation functions to improve deep neural networks,'' \emph{CoRR}, vol.
	abs/1412.6830, 2014.
	
	\bibitem{xu2015empirical}
	B.~Xu, N.~Wang, T.~Chen, and M.~Li, ``{Empirical Evaluation of Rectified
		Activations in Convolutional Network},'' 2015.
	
	\bibitem{Andrej42455}
	A.~Karpathy, G.~Toderici, S.~Shetty, T.~Leung, R.~Sukthankar, and L.~Fei-Fei,
	``Large-scale video classification with convolutional neural networks,'' in
	\emph{Proceedings of International Computer Vision and Pattern Recognition
		(CVPR 2014)}, 2014.
	
	\bibitem{NIPS2012Alex}
	A.~Krizhevsky, I.~Sutskever, and G.~E. Hinton, ``Imagenet classification with
	deep convolutional neural networks,'' in \emph{Advances in Neural Information
		Processing Systems 25}, F.~Pereira, C.~J.~C. Burges, L.~Bottou, and K.~Q.
	Weinberger, Eds.\hskip 1em plus 0.5em minus 0.4em\relax Curran Associates,
	Inc., 2012, pp. 1097--1105.
	
	\bibitem{Simonyan15}
	K.~Simonyan and A.~Zisserman, ``Very deep convolutional networks for
	large-scale image recognition,'' in \emph{International Conference on
		Learning Representations}, 2015.
	
	\bibitem{He7780459}
	K.~{He}, X.~{Zhang}, S.~{Ren}, and J.~{Sun}, ``Deep residual learning for image
	recognition,'' in \emph{2016 IEEE Conference on Computer Vision and Pattern
		Recognition (CVPR)}, June 2016, pp. 770--778.
	
	\bibitem{Jordan2018}
	J.~{Jordan}, ``Common architectures in convolutional neural networks,'' in
	\emph{https://www.jeremyjordan.me/convnet-architectures/}, 2018.
	
	\bibitem{lipton2015}
	Z.~C. Lipton, J.~Berkowitz, and C.~Elkan, ``A critical review of recurrent
	neural networks for sequence learning,'' 2015.
	
	\bibitem{WittenFrankHall11}
	I.~H. Witten, E.~Frank, and M.~A. Hall, \emph{Data Mining: Practical Machine
		Learning Tools and Techniques}, 3rd~ed., ser. Morgan Kaufmann Series in Data
	Management Systems.\hskip 1em plus 0.5em minus 0.4em\relax Amsterdam: Morgan
	Kaufmann, 2011.
	
	\bibitem{HochSchm97}
	S.~Hochreiter and J.~Schmidhuber, ``Long short-term memory,'' \emph{Neural
		Computation}, vol.~9, no.~8, pp. 1735--1780, 1997.
	
	\bibitem{chung2014empirical}
	J.~Chung, C.~Gulcehre, K.~Cho, and Y.~Bengio, ``Empirical evaluation of gated
	recurrent neural networks on sequence modeling,'' 2014.
	
	\bibitem{Junyoung2014}
	C.~Junyoung, C.~Gulcehre, K.~Cho, and Y.~Bengio, ``\BIBforeignlanguage{English
		(US)}{Empirical evaluation of gated recurrent neural networks on sequence
		modeling},'' in \emph{\BIBforeignlanguage{English (US)}{NIPS 2014 Workshop on
			Deep Learning, December 2014}}, 2014.
	
	\bibitem{NIPS2014_Goodfellow}
	I.~Goodfellow, J.~Pouget-Abadie, M.~Mirza, B.~Xu, D.~Warde-Farley, S.~Ozair,
	A.~Courville, and Y.~Bengio, ``Generative adversarial nets,'' in
	\emph{Advances in Neural Information Processing Systems 27}, Z.~Ghahramani,
	M.~Welling, C.~Cortes, N.~D. Lawrence, and K.~Q. Weinberger, Eds.\hskip 1em
	plus 0.5em minus 0.4em\relax Curran Associates, Inc., 2014, pp. 2672--2680.
	
	\bibitem{Jaturapitpornchai_2019}
	R.~Jaturapitpornchai, M.~Matsuoka, N.~Kanemoto, S.~Kuzuoka, R.~Ito, and
	R.~Nakamura, ``Newly built construction detection in sar images using deep
	learning,'' \emph{Remote Sensing}, vol.~11, no.~12, pp. 1--24, Jun 2019.
	
	\bibitem{Yang20199}
	M.~{Yang}, L.~{Jiao}, F.~{Liu}, B.~{Hou}, and S.~{Yang}, ``Transferred deep
	learning-based change detection in remote sensing images,'' \emph{IEEE
		Transactions on Geoscience and Remote Sensing}, vol.~57, no.~9, pp.
	6960--6973, Sep. 2019.
	
	\bibitem{7120131}
	M.~{Gong}, J.~{Zhao}, J.~{Liu}, Q.~{Miao}, and L.~{Jiao}, ``Change detection in
	synthetic aperture radar images based on deep neural networks,'' \emph{IEEE
		Transactions on Neural Networks and Learning Systems}, vol.~27, no.~1, pp.
	125--138, 2016.
	
	\bibitem{Lyu_2016}
	H.~Lyu, H.~Lu, and L.~Mou, ``Learning a transferable change rule from a
	recurrent neural network for land cover change detection,'' \emph{Remote
		Sensing}, vol.~8, no.~6, p. 506, Jun 2016.
	
	\bibitem{ZHANG201624}
	P.~Zhang, M.~Gong, L.~Su, J.~Liu, and Z.~Li, ``Change detection based on deep
	feature representation and mapping transformation for
	multi-spatial-resolution remote sensing images,'' \emph{ISPRS Journal of
		Photogrammetry and Remote Sensing}, vol. 116, pp. 24 -- 41, 2016.
	
	\bibitem{Zhao2007SpectralFS}
	Z.~Zhao and H.~Liu, ``Spectral feature selection for supervised and
	unsupervised learning,'' in \emph{ICML '07}, 2007.
	
	\bibitem{chung-etal-2018-supervised}
	Y.-A. Chung, H.-Y. Lee, and J.~Glass, ``Supervised and unsupervised transfer
	learning for question answering,'' in \emph{Proceedings of the 2018
		Conference of the North {A}merican Chapter of the Association for
		Computational Linguistics: Human Language Technologies}, vol.~1.\hskip 1em
	plus 0.5em minus 0.4em\relax New Orleans, Louisiana: Association for
	Computational Linguistics, Jun. 2018, pp. 1585--1594.
	
	\bibitem{Bengio2013}
	Y.~Bengio, ``Deep learning of representations: Looking forward,'' in
	\emph{Statistical Language and Speech Processing}, A.-H. Dediu,
	C.~Mart{\'i}n-Vide, R.~Mitkov, and B.~Truthe, Eds.\hskip 1em plus 0.5em minus
	0.4em\relax Berlin, Heidelberg: Springer Berlin Heidelberg, 2013, pp. 1--37.
	
	\bibitem{Penatti7301382}
	O.~A.~B. {Penatti}, K.~{Nogueira}, and J.~A. {dos Santos}, ``Do deep features
	generalize from everyday objects to remote sensing and aerial scenes
	domains?'' in \emph{2015 IEEE Conference on Computer Vision and Pattern
		Recognition Workshops (CVPRW)}, June 2015, pp. 44--51.
	
	\bibitem{Ronneberger2015}
	O.~Ronneberger, P.~Fischer, and T.~Brox, ``U-net: Convolutional networks for
	biomedical image segmentation,'' in \emph{Medical Image Computing and
		Computer-Assisted Intervention -- MICCAI 2015}, N.~Navab, J.~Hornegger, W.~M.
	Wells, and A.~F. Frangi, Eds.\hskip 1em plus 0.5em minus 0.4em\relax Cham:
	Springer International Publishing, 2015, pp. 234--241.
	
	\bibitem{Mahmoud2019}
	Z.~M. Hamdi, M.~Brandmeier, and C.~Straub, ``{Forest Damage Assessment Using
		Deep Learning on High Resolution Remote Sensing Data},'' \emph{{Remote
			Sensing}}, vol.~{11}, no.~{17}, {SEP 1} {2019}.
	
	\bibitem{Peng2019}
	D.~Peng, Y.~Zhang, and H.~Guan, ``{End-to-End Change Detection for High
		Resolution Satellite Images Using Improved UNet plus},'' \emph{{Remote
			Sensing}}, vol.~{11}, no.~{11}, {JUN 1} {2019}.
	
	\bibitem{Wang2019}
	Q.~Wang, Z.~Yuan, Q.~Du, and X.~Li, ``{GETNET: A General End-to-End 2-D CNN
		Framework for Hyperspectral Image Change Detection},'' \emph{{IEEE
			Transactions on Geoscience and Remote Sensing }}, vol.~{57}, no.~{1}, pp.
	{3--13}, {JAN} {2019}.
	
	\bibitem{Malila1980}
	W.~A. Malila, ``Change vector analysis: An approach for detecting forest
	changes with landsat,'' in \emph{LARS Symp.}, 1980.
	
	\bibitem{Wahyu2018}
	W.~Wiratama, J.~Lee, S.-E. Park, and D.~Sim, ``{Dual-Dense Convolution Network
		for Change Detection of High-Resolution Panchromatic Imagery},''
	\emph{{Applied Sciences-Basel}}, vol.~{8}, no.~{10}, {OCT} {2018}.
	
	\bibitem{ZhangChi2019}
	C.~Zhang, S.~Wei, S.~Ji, and M.~Lu, ``{Detecting Large-Scale Urban Land Cover
		Changes from Very High Resolution Remote Sensing Images Using CNN-Based
		Classification},'' \emph{{ISPRS International Journal of Geo-Information}},
	vol.~{8}, no.~{4}, {APR} {2019}.
	
	\bibitem{Daudt2019}
	R.~C. Daudt, B.~L. Saux, A.~Boulch, and Y.~Gousseau, ``Multitask learning for
	large-scale semantic change detection,'' \emph{Computer Vision and Image
		Understanding}, vol. 187, p. 102783, 2019.
	
	\bibitem{Zhang20199}
	W.~Zhang and X.~Lu, ``{The Spectral-Spatial Joint Learning for Change Detection
		in Multispectral Imagery},'' \emph{{Remote Sensing}}, vol.~{11}, no.~{3},
	{FEB 1} {2019}.
	
	\bibitem{ZhenchaoZhang2018}
	Z.~Zhang, G.~Vosselman, M.~Gerke, D.~Tuia, and M.~Y. Yang, ``Change detection
	between multimodal remote sensing data using siamese {CNN},'' \emph{CoRR},
	vol. abs/1807.09562, 2018.
	
	\bibitem{Cao2019a}
	X.~{Cao}, Y.~{Ji}, L.~{Wang}, B.~{Ji}, L.~{Jiao}, and J.~{Han}, ``Sar image
	change detection based on deep denoising and cnn,'' \emph{IET Image
		Processing}, vol.~13, no.~9, pp. 1509--1515, 2019.
	
	\bibitem{Wahyu20199}
	W.~Wiratama and D.~Sim, ``{Fusion Network for Change Detection of
		High-Resolution Panchromatic Imagery},'' \emph{{Applied Sciences-Basel}},
	vol.~{9}, no.~{7}, {APR 1} {2019}.
	
	\bibitem{Cao2019b}
	C.~Cao, S.~Dragicevic, and S.~Li, ``Land-use change detection with
	convolutional neural network methods,'' \emph{Environments}, vol.~6, 2019.
	
	\bibitem{Lv2018}
	N.~{Lv}, C.~{Chen}, T.~{Qiu}, and A.~K. {Sangaiah}, ``Deep learning and
	superpixel feature extraction based on contractive autoencoder for change
	detection in sar images,'' \emph{IEEE Transactions on Industrial
		Informatics}, vol.~14, no.~12, pp. 5530--5538, Dec 2018.
	
	\bibitem{Zhangh2016}
	H.~{Zhang}, M.~{Gong}, P.~{Zhang}, L.~{Su}, and J.~{Shi}, ``Feature-level
	change detection using deep representation and feature change analysis for
	multispectral imagery,'' \emph{IEEE Geoscience and Remote Sensing Letters},
	vol.~13, no.~11, pp. 1666--1670, Nov 2016.
	
	\bibitem{Su2017}
	L.~Su, M.~Gong, P.~Zhang, M.~Zhang, J.~Liu, and H.~Yang, ``Deep learning and
	mapping based ternary change detection for information unbalanced images,''
	\emph{Pattern Recognition}, vol.~66, pp. 213 -- 228, 2017.
	
	\bibitem{Feng2017}
	F.~Gao, X.~Liu, J.~Dong, G.~Zhong, and M.~Jian, ``Change detection in sar
	images based on deep semi-nmf and svd networks,'' \emph{Remote Sensing},
	vol.~9, p. 435, 05 2017.
	
	\bibitem{Wang6165290}
	Y.~{Wang} and Y.~{Zhang}, ``Nonnegative matrix factorization: A comprehensive
	review,'' \emph{IEEE Transactions on Knowledge and Data Engineering},
	vol.~25, no.~6, pp. 1336--1353, 2013.
	
	\bibitem{XueLG13}
	J.~Xue, J.~Li, and Y.~Gong, ``Restructuring of deep neural network acoustic
	models with singular value decomposition.'' in \emph{INTERSPEECH}, F.~Bimbot,
	C.~Cerisara, C.~Fougeron, G.~Gravier, L.~Lamel, F.~Pellegrino, and
	P.~Perrier, Eds.\hskip 1em plus 0.5em minus 0.4em\relax ISCA, 2013, pp.
	2365--2369.
	
	\bibitem{Gongg2019}
	M.~{Gong}, Y.~{Yang}, T.~{Zhan}, X.~{Niu}, and S.~{Li}, ``A generative
	discriminatory classified network for change detection in multispectral
	imagery,'' \emph{IEEE Journal of Selected Topics in Applied Earth
		Observations and Remote Sensing}, vol.~12, no.~1, pp. 321--333, Jan 2019.
	
	\bibitem{Geng2019}
	J.~{Geng}, X.~{Ma}, X.~{Zhou}, and H.~{Wang}, ``Saliency-guided deep neural
	networks for sar image change detection,'' \emph{IEEE Transactions on
		Geoscience and Remote Sensing}, vol.~57, no.~10, pp. 7365--7377, 2019.
	
	\bibitem{Geva19999}
	A.~B. Geva, ``Hierarchical unsupervised fuzzy clustering,'' \emph{IEEE
		Transactions on Fuzzy Systems}, vol.~7, no.~6, pp. 723--733, 1999.
	
	\bibitem{Xuelong2019}
	X.~Li, Z.~Yuan, and Q.~Wang, ``Unsupervised deep noise modeling for
	hyperspectral image change detection,'' \emph{Remote Sensing}, vol.~11,
	no.~3, 2019.
	
	\bibitem{Huang2019}
	F.~Huang, Y.~Yu, and T.~Feng, ``Automatic building change image quality
	assessment in high resolution remote sensing based on deep learning,''
	\emph{Journal of Visual Communication and Image Representation}, vol.~63, p.
	102585, 2019.
	
	\bibitem{SURESH2009541}
	S.~Suresh, R.~V. Babu, and H.~J. Kim, ``No-reference image quality assessment
	using modified extreme learning machine classifier,'' \emph{Applied Soft
		Computing}, vol.~9, no.~2, pp. 541 -- 552, 2009.
	
	\bibitem{Pan2010}
	S.~J. {Pan} and Q.~{Yang}, ``A survey on transfer learning,'' \emph{IEEE
		Transactions on Knowledge and Data Engineering}, vol.~22, no.~10, pp.
	1345--1359, 2010.
	
	\bibitem{Weiss2016}
	K.~Weiss, T.~M. Khoshgoftaar, and D.~Wang, ``A survey of transfer learning,''
	\emph{Journal of Big Data}, vol.~3, 2016.
	
	\bibitem{zhao2018object}
	Z.-Q. Zhao, P.~Zheng, S.-t. Xu, and X.~Wu, ``Object detection with deep
	learning: A review,'' 2018.
	
	\bibitem{ABDALLA2019105091}
	A.~Abdalla, H.~Cen, L.~Wan, R.~Rashid, H.~Weng, W.~Zhou, and Y.~He,
	``Fine-tuning convolutional neural network with transfer learning for
	semantic segmentation of ground-level oilseed rape images in a field with
	high weed pressure,'' \emph{Computers and Electronics in Agriculture}, vol.
	167, p. 105091, 2019.
	
	\bibitem{Bin2017}
	B.~Hou, Y.~Wang, and Q.~Liu, ``Change detection based on deep features and low
	rank,'' \emph{IEEE Geoscience and Remote Sensing Letters}, vol.~PP, pp. 1--5,
	11 2017.
	
	\bibitem{imagenet_cvpr09}
	J.~Deng, W.~Dong, R.~Socher, L.-J. Li, K.~Li, and L.~Fei-Fei, ``{ImageNet: A
		Large-Scale Hierarchical Image Database},'' in \emph{CVPR09}, 2009.
	
	\bibitem{7907303}
	G.~{Xia}, J.~{Hu}, F.~{Hu}, B.~{Shi}, X.~{Bai}, Y.~{Zhong}, L.~{Zhang}, and
	X.~{Lu}, ``Aid: A benchmark data set for performance evaluation of aerial
	scene classification,'' \emph{IEEE Transactions on Geoscience and Remote
		Sensing}, vol.~55, no.~7, pp. 3965--3981, 2017.
	
	\bibitem{Venugopal2019}
	N.~Venugopal, ``Sample selection based change detection with dilated network
	learning in remote sensing images,'' \emph{Sensing and Imaging}, vol.~20, 12
	2019.
	
	\bibitem{Kaiming2015}
	K.~He, X.~Zhang, S.~Ren, and J.~Sun, ``Deep residual learning for image
	recognition,'' \emph{CoRR}, vol. abs/1512.03385, 2015.
	
	\bibitem{ZhangM2019}
	M.~{Zhang}, G.~{Xu}, K.~{Chen}, M.~{Yan}, and X.~{Sun}, ``Triplet-based
	semantic relation learning for aerial remote sensing image change
	detection,'' \emph{IEEE Geoscience and Remote Sensing Letters}, vol.~16,
	no.~2, pp. 266--270, Feb 2019.
	
	\bibitem{Liang}
	L.~Chen, G.~Papandreou, I.~Kokkinos, K.~Murphy, and A.~L. Yuille, ``Deeplab:
	Semantic image segmentation with deep convolutional nets, atrous convolution,
	and fully connected crfs,'' \emph{CoRR}, vol. abs/1606.00915, 2016.
	
	\bibitem{pascal-voc-2012}
	M.~Everingham, L.~Van~Gool, C.~K.~I. Williams, J.~Winn, and A.~Zisserman, ``The
	pascal visual object classes (voc) challenge,'' \emph{International Journal
		of Computer Vision}, vol.~88, no.~2, pp. 303--338, 2010.
	
	\bibitem{Fang2019DualLS}
	B.~Fang, L.~Pan, and R.~Kou, ``Dual learning-based siamese framework for change
	detection using bi-temporal vhr optical remote sensing images,'' \emph{Remote
		Sensing}, vol.~11, p. 1292, 2019.
	
	\bibitem{Gao2019}
	Y.~{Gao}, F.~{Gao}, J.~{Dong}, and S.~{Wang}, ``Transferred deep learning for
	sea ice change detection from synthetic-aperture radar images,'' \emph{IEEE
		Geoscience and Remote Sensing Letters}, vol.~16, no.~10, pp. 1655--1659, Oct
	2019.
	
	\bibitem{Mnih2015}
	V.~Mnih, K.~Kavukcuoglu, D.~Silver, A.~A. Rusu, J.~Veness, M.~G. Bellemare,
	A.~Graves, M.~Riedmiller, A.~K. Fidjeland, G.~Ostrovski, S.~Petersen,
	C.~Beattie, A.~Sadik, I.~Antonoglou, H.~King, D.~Kumaran, D.~Wierstra,
	S.~Legg, and D.~Hassabis, ``Human-level control through deep reinforcement
	learning,'' \emph{Nature}, vol. 518, pp. 529--533, 2015.
	
	\bibitem{7989381}
	Y.~{Zhu}, R.~{Mottaghi}, E.~{Kolve}, J.~J. {Lim}, A.~{Gupta}, L.~{Fei-Fei}, and
	A.~{Farhadi}, ``Target-driven visual navigation in indoor scenes using deep
	reinforcement learning,'' in \emph{2017 IEEE International Conference on
		Robotics and Automation (ICRA)}, 2017, pp. 3357--3364.
	
	\bibitem{9043893}
	T.~T. {Nguyen}, N.~D. {Nguyen}, and S.~{Nahavandi}, ``Deep reinforcement
	learning for multiagent systems: A review of challenges, solutions, and
	applications,'' \emph{IEEE Transactions on Cybernetics}, pp. 1--14, 2020.
	
	\bibitem{8911507}
	C.~{Hoel}, K.~{Driggs-Campbell}, K.~{Wolff}, L.~{Laine}, and M.~{Kochenderfer},
	``Combining planning and deep reinforcement learning in tactical decision
	making for autonomous driving,'' \emph{IEEE Transactions on Intelligent
		Vehicles}, pp. 1--12, 2019.
	
	\bibitem{8998330}
	Y.~{Dai}, D.~{Xu}, K.~{Zhang}, S.~{Maharjan}, and Y.~{Zhang}, ``Deep
	reinforcement learning and permissioned blockchain for content caching in
	vehicular edge computing and networks,'' \emph{IEEE Transactions on Vehicular
		Technology}, vol.~69, no.~4, pp. 4312--4324, 2020.
	
	\bibitem{TENG2020107188}
	Z.~Teng, B.~Zhang, and J.~Fan, ``Three-step action search networks with deep
	q-learning for real-time object tracking,'' \emph{Pattern Recognition}, vol.
	101, pp. 1--11, 2020.
	
	\bibitem{8642452}
	W.~{Luo}, P.~{Sun}, F.~{Zhong}, W.~{Liu}, T.~{Zhang}, and Y.~{Wang},
	``End-to-end active object tracking and its real-world deployment via
	reinforcement learning,'' \emph{IEEE Transactions on Pattern Analysis and
		Machine Intelligence}, pp. 1--14, 2019.
	
	\bibitem{Liu_2019_ICCV}
	Z.~Liu, J.~Wang, S.~Gong, H.~Lu, and D.~Tao, ``Deep reinforcement active
	learning for human-in-the-loop person re-identification,'' in \emph{The IEEE
		International Conference on Computer Vision (ICCV)}, October 2019.
	
	\bibitem{8666162}
	W.~{Zhang}, X.~{He}, W.~{Lu}, H.~{Qiao}, and Y.~{Li}, ``Feature aggregation
	with reinforcement learning for video-based person re-identification,''
	\emph{IEEE Transactions on Neural Networks and Learning Systems}, vol.~30,
	no.~12, pp. 3847--3852, 2019.
	
	\bibitem{8694781}
	A.~{Shrestha} and A.~{Mahmood}, ``Review of deep learning algorithms and
	architectures,'' \emph{IEEE Access}, vol.~7, pp. 53\,040--53\,065, 2019.
	
	\bibitem{Volodymyr2016}
	V.~Mnih, A.~P. Badia, M.~Mirza, A.~Graves, T.~Lillicrap, T.~Harley, D.~Silver,
	and K.~Kavukcuoglu, ``Asynchronous methods for deep reinforcement learning,''
	in \emph{Proceedings of The 33rd International Conference on Machine
		Learning}, ser. Proceedings of Machine Learning Research, M.~F. Balcan and
	K.~Q. Weinberger, Eds., vol.~48.\hskip 1em plus 0.5em minus 0.4em\relax New
	York, New York, USA: PMLR, 20--22 Jun 2016, pp. 1928--1937.
	
	\bibitem{Fu2018ASR}
	K.~Fu, Y.~D. Li, H.~Sun, X.~Yang, G.~Xu, Y.~Li, and X.~Sun, ``A ship rotation
	detection model in remote sensing images based on feature fusion pyramid
	network and deep reinforcement learning,'' \emph{Remote Sensing}, vol.~10,
	pp. 1--26, 2018.
	
	\bibitem{Li2018AnAD}
	Y.~Li, K.~Fu, H.~Sun, and X.~Sun, ``An aircraft detection framework based on
	reinforcement learning and convolutional neural networks in remote sensing
	images,'' \emph{Remote Sensing}, vol.~10, pp. 1--17, 2018.
	
	\bibitem{PENG1994226}
	J.~Peng and R.~J. Williams, ``Incremental multi-step q-learning,'' in
	\emph{Machine Learning Proceedings 1994}, W.~W. Cohen and H.~Hirsh,
	Eds.\hskip 1em plus 0.5em minus 0.4em\relax San Francisco (CA): Morgan
	Kaufmann, 1994, pp. 226 -- 232.
	
	\bibitem{Zhou2017}
	Z.-H. Zhou, ``{A brief introduction to weakly supervised learning},''
	\emph{National Science Review}, vol.~5, no.~1, pp. 44--53, 08 2017.
	
	\bibitem{CHENG201611}
	G.~Cheng and J.~Han, ``A survey on object detection in optical remote sensing
	images,'' \emph{ISPRS Journal of Photogrammetry and Remote Sensing}, vol.
	117, pp. 11 -- 28, 2016.
	
	\bibitem{Li8735810}
	Y.~{Li}, L.~{Guo}, and Z.~{Zhou}, ``Towards safe weakly supervised learning,''
	\emph{IEEE Transactions on Pattern Analysis and Machine Intelligence}, pp.
	1--14, 2019.
	
	\bibitem{Zhou2016WeaklyST}
	P.~Zhou, G.~Cheng, Z.~Liu, S.~Bu, and X.~Hu, ``Weakly supervised target
	detection in remote sensing images based on transferred deep features and
	negative bootstrapping,'' \emph{Multidimensional Systems and Signal
		Processing}, vol.~27, pp. 925--944, 2016.
	
	\bibitem{7153907}
	P.~{Zhou}, D.~{Zhang}, G.~{Cheng}, and J.~{Han}, ``Negative bootstrapping for
	weakly supervised target detection in remote sensing images,'' in \emph{2015
		IEEE International Conference on Multimedia Big Data}, 2015, pp. 318--323.
	
	\bibitem{Jiang20119}
	H.~Jiang, X.~Hu, K.~Li, J.~Zhang, J.~Gong, and M.~Zhang, ``Pga-siamnet: Pyramid
	feature-based attention-guided siamese network for remote sensing
	orthoimagery building change detection,'' \emph{Remote Sensing}, vol.~12,
	no.~3, 2020.
	
\end{thebibliography}
\end{document}